\documentclass[10pt,twocolumn,letterpaper]{article}

\usepackage[pagenumbers]{cvpr} % To force page numbers, e.g. for an arXiv version

\usepackage{url}
\usepackage{amsmath}
\usepackage{amsfonts}
\usepackage{amssymb}

\usepackage{graphicx}
\usepackage{subcaption}
\usepackage{float}

\usepackage{booktabs}
\usepackage{multirow}
\usepackage{tabularx}

\usepackage{microtype}
\usepackage{nicefrac}

\usepackage{xcolor}
\usepackage{pifont}

\usepackage{enumitem}

\usepackage[most]{tcolorbox}
\tcbuselibrary{listings,breakable,skins}
\newtcblisting{promptbox}[1]{
    enhanced,
    breakable,
    listing only,
    colback=gray!5,
    colframe=black!85,
    colbacktitle=black!80,
    coltitle=white,
    title={#1},
    fonttitle=\bfseries\small,
    boxrule=0.6pt,
    arc=1.5mm,
    left=2mm,
    right=2mm,
    top=1.5mm,
    bottom=1.5mm,
    before skip=5pt,
    after skip=7pt,
    listing options={
    basicstyle=\small\rmfamily,
    breaklines=true,
    breakatwhitespace=false,
    breakindent=0pt,
    breakautoindent=false,
    columns=fullflexible,
    keepspaces=true,
    showstringspaces=false
    }
}

\definecolor{cvprblue}{rgb}{0.21,0.49,0.74}
\usepackage[pagebackref,breaklinks,colorlinks,allcolors=cvprblue]{hyperref}

\def\confName{CVPR}
\def\confYear{2027}

\title{FAIRP: Fidelity-Aware Image-only Retrieval Poisoning in Multimodal RAG}

\author{%
{\bfseries
Shuo Ju$^{1,2}$,
Huashan Chen$^{1,2}$,
Haotang Li$^{3}$,
Kebin Peng$^{4}$,
Feng Liu$^{1,2}$,
Sen He$^{3}$,
Han Xu$^{3}$
}\\[0.3cm]
{\normalfont $^{1}$ Institute of Information Engineering} \\
{\normalfont $^{2}$ University of the Chinese Academy of Sciences} \\
{\normalfont $^{3}$ University of Arizona} \\
{\normalfont $^{4}$ East Carolina University}
}
\begin{document}

\maketitle

\begin{abstract}
Existing knowledge base (KB) poisoning attacks on multimodal retrieval-augmented generation (RAG) typically rely on attacker-crafted misleading text or images designed around predefined outputs. We propose FAIRP (Fidelity-Aware Image-only Retrieval Poisoning), an image-only poisoning framework that manipulates retrieval without controlling paired text, specifying target answers, or using gradients from the downstream vision-language model (VLM). Starting from images unrelated to the query, the attacker optimizes pixel perturbations to improve their retrieval rankings; after submission, the KB automatically generates paired text through the normal ingestion process. The resulting image-text records compete with existing records for retrieval, alter evidence selection, and turn correct Clean-RAG answers into incorrect ones, while the original KB records and all model parameters remain unchanged. We evaluate our method on InfoSeek using three retrievers under three retrieval configurations and four VLMs, achieving average ASRs of 77.6\% and 51.6\% for fixed-query and category-shared attacks, respectively. Experiments with two models, each serving as both the retriever and captioner, demonstrate that our attack can manipulate retrieval while preserving caption fidelity, revealing differences in task-specific visual feature utilization. Further analyses examine attack effectiveness through question reformulation, transferability, and ablation studies.
\end{abstract}
\section{Introduction}

Retrieval-augmented generation (RAG) provides models with additional knowledge by retrieving relevant evidence from external knowledge bases (KB)~\cite{lewis2020retrieval, guu2020retrieval, izacard2021leveraging}. In multimodal question answering, this mechanism incorporates visual and textual evidence to support questions that are difficult to answer using the user input and the model's parametric knowledge alone~\cite{chen2022murag, hu2023reveal, hu2025mrag}. For KBs containing image-text records, a retriever selects relevant records according to the user query and supplies the retrieved images and their paired text to a vision-language model (VLM) for answer generation. These records therefore directly influence the external evidence available to the VLM and its final responses. As new records are added to the KB, they may provide useful knowledge but can also alter retrieval results, potentially displacing relevant evidence with irrelevant records and affecting VLM answers. This makes poisoning attack a relevant security concern for multimodal RAG.

Existing RAG poisoning attacks manipulate textual records~\cite{zou2025poisonedrag, gong2025topic, liu2026warp, shafran2025machine} or inject image-text pairs~\cite{ha-etal-2026-mm, yang2026knowledge, liu2025poisoned, zhang2025poisonedeye} to influence RAG. Recent attacks~\cite{shereen2026one, liang2026vis} further demonstrate this threat through joint retrieval-generation optimization or semantic image editing. Despite their different mechanisms, these methods generally rely on either attacker-crafted malicious text or predefined target answers. However, an important attack surface has been overlooked: \textit{image-only poisoning can manipulate retrieval rankings and introduce irrelevant records into the evidence, even when their paired captions are generated by the KB and remain beyond the attacker's control, revealing that image-caption consistency alone does not guarantee retrieval relevance.}

\begin{figure}[t]
    \centering
    \includegraphics[width=\linewidth]{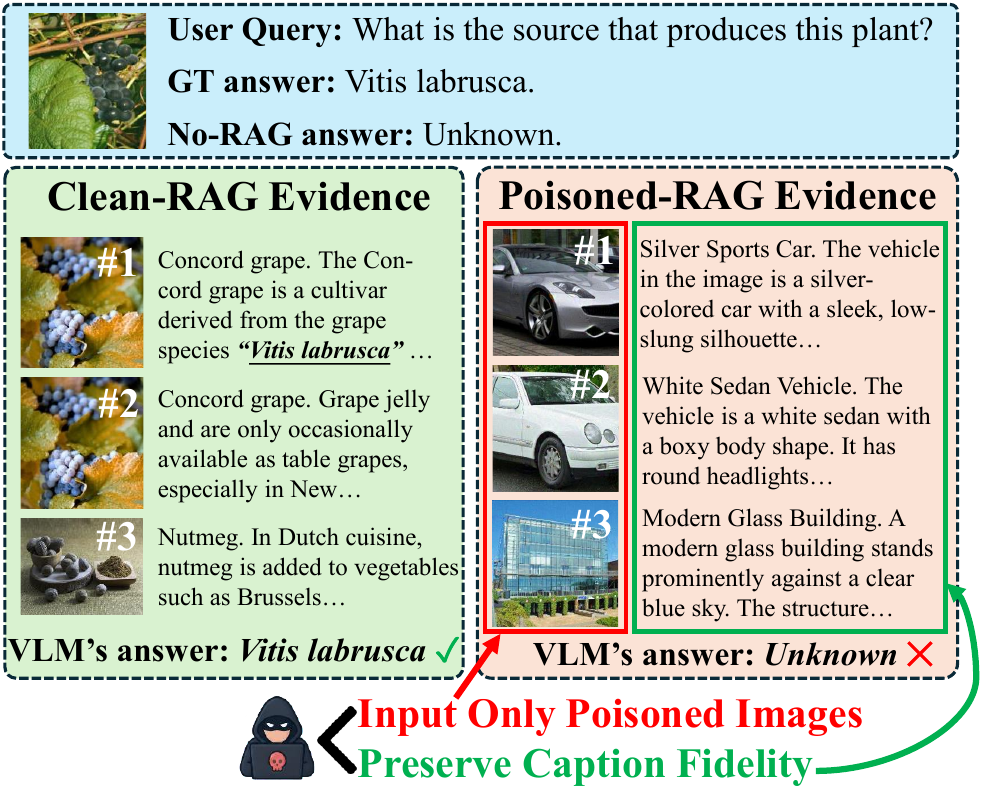}
    \caption{Attack demonstration: the attacker inputs only poisoned images, whose paired captions are generated through the KB's normal ingestion process and remain correct to the images, yet the resulting records hijack RAG by displacing original evidence and causing incorrect VLM responses. No-RAG fails to answer, Clean-RAG correctly answers \textit{Vitis labrusca}, whereas Poisoned-RAG returns \textit{Unknown} after the original evidence is displaced.}
    \label{fig:intro}
\end{figure}

This motivates us to explore a novel \textbf{image-only poisoning attack: } optimizing perturbations on poisoned images to improve their retrieval rankings while preserving generated caption fidelity, without predefining any target answer or altering any text.
An attack demonstration is illustrated in Fig.~\ref{fig:intro}. The attacker only needs to submit poisoned images to the KB, where its captioning model generates paired captions and stores the resulting image-text pairs as new records. Although the poisoned images can manipulate retrieval rankings, they still remain sufficiently preserved for the KB's captioning model to generate correct captions. Consequently, the new records \textit{seem harmless} but may displace original evidence. The VLM thus receives the irrelevant evidence and generates an incorrect answer to the query. Importantly, our attack only requires submitting poisoned images to manipulate retrieval; that is, we do not need to craft malicious text or predefine target answers.

However, realizing this idea remains challenging. First, the attacker can optimize only images, whereas newly ingested KB records also contain generated text, which may weaken the advantage achieved through image optimization. Second, perturbations may alter the original visual semantics, causing generated captions to deviate from the image content and making the attack detectable through human inspection. Third, changes in wording may also alter query representations and affect the retrieval of poisoned records. Fourth, even when poisoned records are retrieved, the VLM may ignore irrelevant content or answer correctly using the remaining evidence. \textit{To address these challenges,} we propose FAIRP, a fidelity-aware image-only retrieval poisoning method that integrates Image-only Retrieval Optimization (Sec.~\ref{sec:image_optimization}), a Caption Preservation Objective, and visual regularization (Sec.~\ref{sec:optimization_objective}). FAIRP supports both Fixed-query and Category-shared attacks (Sec.~\ref{sec:image_optimization}), with its robustness to question reformulation further evaluated in Sec.~\ref{sec:question_reformulation}. Finally, we deploy the complete RAG pipeline, including VLM answer generation, to evaluate whether FAIRP can turn correct Clean-RAG responses into incorrect ones (Sec.~\ref{sec:main_results}).

We evaluate our method on InfoSeek using three retrievers, three retrieval configurations, and four VLMs. Under multimodal RAG with equal-weight image-text fusion, fixed-query and category-shared attacks achieve average ASRs of 77.6\% and 51.6\%, respectively. For fixed-query attacks, 84.4\% of originally successful cases remain successful after question reformulation. These results demonstrate that images optimized solely against the retriever can enter the retrieved evidence and turn correct Clean-RAG answers into incorrect ones. Experiments with models serving as both retriever and captioner demonstrate that our attack can manipulate retrieval while minimizing its impact on caption generation. We further examine attack effectiveness through ablation study on the number of poisoned images and image-text fusion weights, as well as transferability.

\textbf{Our contributions:} (1) We reveal that image-caption consistency does not guarantee retrieval relevance, as image-only poisoning can displace useful evidence with irrelevant but correct captions in multimodal RAG. (2) We propose FAIRP, an image-only poisoning method that manipulates retrieval rankings and induces incorrect VLM responses while preserving caption fidelity, without controlling paired text or predefining target answers. (3) We evaluate FAIRP on InfoSeek with three retrievers, three retrieval configurations, and four VLMs, achieving average ASRs of 77.6\% and 51.6\% for fixed-query and category-shared attacks.
\section{Related Work}
\label{sec:related_work}
\begin{figure*}[t]
    \centering
    \includegraphics[width=\linewidth]{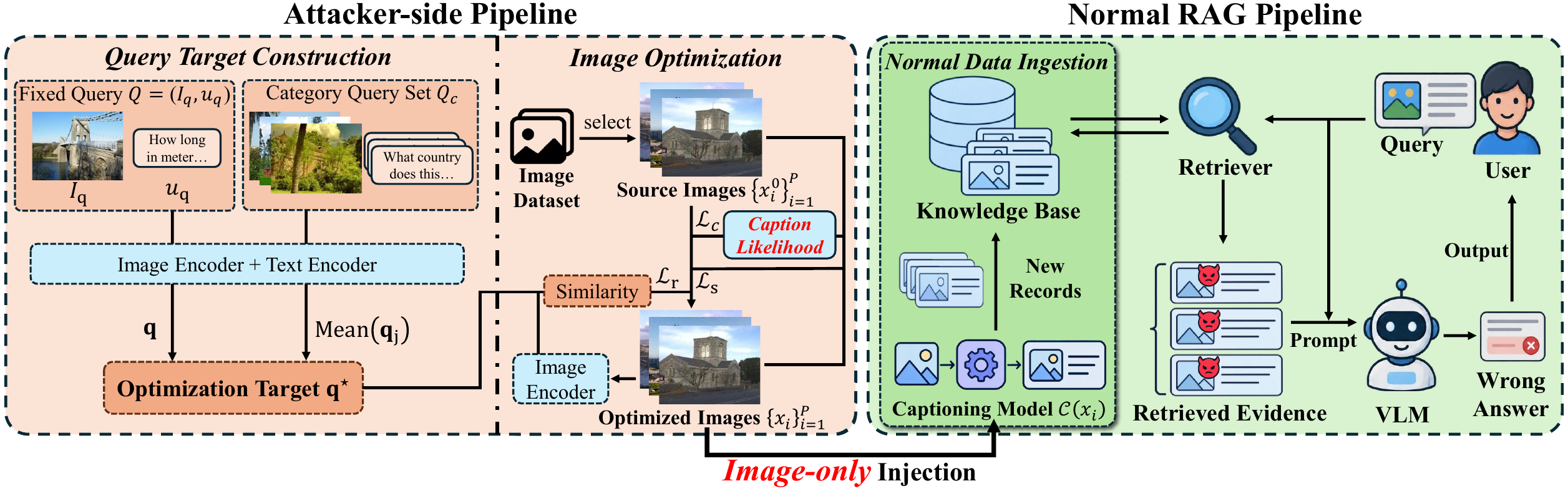}
    \caption{Overview of FAIRP. Query Target Construction establishes optimization targets under fixed-query and category-shared settings. Image Optimization updates source images using retrieval, caption-preservation, and visual-regularization objectives. The optimized images are then submitted to the KB, where the normal ingestion process generates paired captions. The resulting poisoned records can enter the top-$K$ retrieved evidence, altering the VLM's responses.}
    \label{fig:method}
\end{figure*}
\textbf{Multimodal RAG.}
Retrieval-augmented generation supplies language models with external evidence for knowledge-intensive tasks~\cite{lewis2020retrieval}. Dense Passage Retrieval~\cite{karpukhin2020dense} learns query-passage matching, while Fusion-in-Decoder~\cite{izacard2021leveraging} aggregates multiple retrieved passages for answer generation. CLIP and SigLIP align images and text in a shared representation space~\cite{radford2021learning,zhai2023sigmoid}, while BLIP-2~\cite{li2023blip} and LLaVA~\cite{liu2023visual} incorporate visual features into language generation. MuRAG~\cite{chen2022murag} and REVEAL~\cite{hu2023reveal} augment generation with external multimodal memories, and UniIR~\cite{wei2024uniir} unifies retrieval across different query and evidence modalities. Related benchmarks cover complementary reasoning requirements: MultiModalQA~\cite{talmor2021multimodalqa} requires evidence integration across text, tables, and images, while WebQA~\cite{chang2022webqa} and InfoSeek~\cite{chen2023can} focus on multimodal multi-hop reasoning and visual information-seeking questions, respectively. ColPali~\cite{faysse2025colpali} supports retrieval over page images, while VisRAG~\cite{yu2025visrag} integrates visual retrieval and generation to retain page layouts, charts, and their relationships with text.

\begin{table}[t]
\centering
\small
\setlength{\tabcolsep}{2.5pt}
\caption{Comparison with related poisoning methods. The last column specifies the representations matched during retrieval: I = image, T = text, I\&T = image-text fusion, and D = document-page.}
\label{tab:related_work}
\begin{tabularx}{\columnwidth}{@{}Xccc@{}}
\toprule
\textbf{Method} &
\shortstack{\textbf{Image-only}\\\textbf{control}} &
\shortstack{\textbf{No predefined}\\\textbf{target output}} &
\shortstack{\textbf{Query $\rightarrow$}\\\textbf{Evidence}} \\
\midrule
PoisonedEye~\cite{zhang2025poisonedeye} & $\times$ & $\times$ & I\&T $\rightarrow$ I\&T \\
M\textsuperscript{3}Att~\cite{yang2026knowledge} & $\times$ & $\times$ & I $\rightarrow$ I \\
MM-PoisonRAG~\cite{ha-etal-2026-mm} & $\times$ & $\times$ & T $\rightarrow$ I \\
OnePic~\cite{shereen2026one} & $\checkmark$ & $\times$ & T $\rightarrow$ D \\
Vis-Poison~\cite{liang2026vis} & $\checkmark$ & $\times$ & T $\rightarrow$ T/I \\
\textbf{FAIRP (Ours)} & $\checkmark$ & $\checkmark$ & I\&T $\rightarrow$ I\&T \\
\bottomrule
\end{tabularx}
\end{table}

\textbf{RAG Poisoning Attacks.}
Existing textual RAG attacks manipulate retrieved content to alter evidence selection and steer downstream generation toward attacker-desired outputs~\cite{zhong2023poisoning,zou2025poisonedrag,gong2025topic,shafran2025machine,greshake2023not,zhan2024injecagent}. Existing image-text poisoning methods typically optimize or generate images to attract retrieval, then use paired malicious text to inject false knowledge or behavioral instructions into generation. PoisonedEye~\cite{zhang2025poisonedeye} applies this design to fixed-query and category-shared attacks, while MM-PoisonRAG~\cite{ha-etal-2026-mm} investigates local and global poisoning. For medical RAG, M\textsuperscript{3}Att~\cite{yang2026knowledge} combines distribution-guided image optimization with malicious medical reports. Some variants require only retriever access without downstream VLM gradients. Despite differences in scope and access requirements, all these methods incorporate attacker-crafted text. In contrast, our attack optimizes only images to manipulate retrieval rankings, without controlling paired text.

\textbf{Adversarial Attacks and Image Poisoning.}
Universal visual perturbations~\cite{moosavi2017universal} and textual triggers~\cite{wallace2019universal} demonstrate that a shared adversarial payload can affect multiple inputs. Adversarial images can also bypass VLM safety constraints~\cite{qi2024visual} or control generated content and instruction following~\cite{bailey2024runtime}. Unlike these direct-input attacks, poisoned images in a knowledge base must first be retrieved to influence answering. OnePic~\cite{shereen2026one} targets visual document RAG; its gradient-based method jointly optimizes retrieval scores and a generation loss against attacker-specified responses. Vis-Poison~\cite{liang2026vis} instead edits local semantic content in query-relevant images to support specified incorrect answers. Both approaches construct poisoned images around predefined outputs or semantic editing. However, our image-only poisoning can manipulate retrieval rankings, without predefining target answers or semantically editing images.

As summarized in Tab.~\ref{tab:related_work}, FAIRP achieves image-only retrieval manipulation while preserving caption fidelity, without attacker-controlled text, semantic image editing, or predefined target outputs.
\section{Threat Model}
\label{threat_model}

\textbf{System Assumptions.}
We consider a multimodal RAG system built upon an image-text KB. The system uses a retriever to retrieve relevant image-text records from the KB and provides them as external evidence to a downstream VLM for answer generation. For newly injected images, the KB generates paired captions through its normal ingestion process. Our attack targets the retrieval stage: \textit{manipulating the retrieval rankings by image-only poisoning, without controlling their paired text,} while preserving caption generation fidelity. The downstream VLM then naturally produces \textit{incorrect responses} based on the manipulated evidence.

\textbf{Adversary Model.}
We assume that the attacker has access to the target query and white-box access to the image-encoding branch of the target retriever. To preserve caption fidelity, the attacker also has white-box access to the KB's captioning model. The attacker can select several source images from image datasets and generate poisoned images through optimization before injecting them into the KB. The attacker's control is restricted to the images of newly injected records; they cannot author, specify, modify, or tamper with the paired text, which is generated by the KB's captioning model. We do not assume that the attacker has access to the downstream VLM, nor does the attacker modify the original KB records, user queries, reference answers, or any model parameters. Instead, the attack is carried out solely through the KB's normal ingestion process, without tampering with textual content, the downstream VLM, or the RAG system.

\textbf{Attack Goal.}
We aim to turn queries correctly answered by Clean-RAG into incorrectly answered queries under Poisoned-RAG by manipulating the retrieved evidence through image-only poisoning while preserving caption.
\section{Method}
\label{sec:design}

\subsection{Problem Formulation and Design Overview}
\label{sec:problem_formulation}

\textbf{Problem Formulation. }Following the threat model in Sec.~\ref{threat_model}, we formulate the attack as an image optimization problem in a multimodal RAG system. The system is built upon an image-text KB, denoted by $\mathcal{D}=\{(I_j,t_j)\}_{j=1}^{N}$, where $I_j$ and $t_j$ represent the image and its paired text in the $j$-th record, respectively. For each newly added image $I$, the KB's captioning model $\mathcal{C}$ generates a caption $t=\mathcal{C}(I)$, forming a new image-text record $r=(I,t)$ in the KB.
Given a user query $Q=(I_q,u_q)$, consisting of an image $I_q$ and a question $u_q$, let $f_I(\cdot)$ and $f_T(\cdot)$ denote the image and text embedding functions of the retriever, respectively. The fused query embedding is defined as $\mathbf{q}=f_I(I_q)+f_T(u_q)$. Similarly, the fused embedding of the $j$-th KB record is $\mathbf{d}_j=f_I(I_j)+f_T(t_j)$, and their similarity is denoted by $\operatorname{sim}(\mathbf{q},\mathbf{d}_j)$. Given a user query $Q$, the retriever then ranks all KB records according to their similarities and selects the top-$K$ evidence for the downstream VLM.

To manipulate this retrieval process, the attacker selects $P$ source images $\{x_i^0\}_{i=1}^{P}$ unrelated to the target query, and optimizes their pixels to obtain poisoned images $\{x_i\}_{i=1}^{P}$. During optimization, the images $\{x_i\}_{i=1}^{P}$ are encouraged to achieve higher similarity while preserving the source captions $t_i^0=\mathcal{C}(x_i^0)$. After optimization, the KB's captioning model generates a caption $t_i^p=\mathcal{C}(x_i)$ for each poisoned image $x_i$, forming a poisoned record $r_i^p=(x_i,t_i^p)$. The attack aims to rank the poisoned records among the top-$K$ evidence and induce incorrect VLM responses.

\textbf{Attack Design Overview. }Inspired by the different sensitivities of retrieval and caption generation, we develop FAIRP, an image-only poisoning attack that manipulates retrieval while preserving caption fidelity, without predefined target answers. As illustrated in Fig.~\ref{fig:method}, our attack consists of two stages: \textit{Query Target Construction} establishes an optimization target under either the fixed-query or category-shared setting (Sec.~\ref{sec:image_optimization}); \textit{Image Optimization} updates the selected source images under three objectives (Sec.~\ref{sec:image_optimization} and \ref{sec:optimization_objective}). And after injection, the poisoned images undergo the KB's normal captioning and retrieval pipeline. The poisoned records may enter the top-$K$ evidence, altering the information provided to the downstream VLM and inducing incorrect responses.

\subsection{Image-only Retrieval Optimization}
\label{sec:image_optimization}

Optimizing a poisoned image requires an embedding target. To accommodate queries at different levels of granularity, we consider two target construction settings: Fixed-query and Category-shared. In the Fixed-query setting, we directly use the fused query embedding $\mathbf{q}$ defined in Sec.~\ref{sec:problem_formulation} as the optimization target. For the Category-shared setting, we first apply spherical clustering to the fused embeddings in a candidate query pool. The resulting clusters are labeled, filtered, and merged according to the visual semantics of representative images, yielding related query sets $\mathcal{Q}_c$. We then compute the normalized mean of all query embeddings within each set to obtain a shared category center $\mathbf{q}_c=\mathcal{N}\left(\frac{1}{|\mathcal{Q}_c|}\sum_{Q_j\in\mathcal{Q}_c}\mathbf{q}_j\right)$, where $\mathcal N(\cdot)$ denotes $L_2$ normalization. We denote the optimization target in both settings by $\mathbf{q}^{\star}$, corresponding to $\mathbf{q}$ and $\mathbf{q}_c$, respectively.

Given the optimization target, the attacker selects $P$ source images $\{x_i^0\}_{i=1}^{P}$ from an image dataset, which are semantically unrelated to the target query. Since the attacker can only modify the images while their paired captions are generated by $\mathcal C$, we optimize the image embeddings of poisoned images against the fused image-text embedding of the target query. For the $i$-th optimized image $x_i$, its retrieval score is defined as
\begin{equation}
s_i=\operatorname{sim}(\mathbf{q}^{\star},f_I(x_i)),
\label{eq:s_i}
\end{equation}
where $f_I(x_i)$ is the image embedding of the poisoned image.

Since each poisoned image will eventually be retrieved together with its generated caption, we use the top-1 retrieval score from the clean KB as a reference and introduce a margin $\kappa$ to compensate for the effect of caption fusion. 
The top-1 retrieval score in the clean KB is defined differently for the two settings:
\begin{equation}
b^{\star}=
\begin{cases}
\displaystyle
\max_{j=1,\ldots,N}
\operatorname{sim}(\mathbf q,\mathbf d_j),
& \text{Fixed-query},\\[6pt]
\displaystyle
\frac{1}{|\mathcal Q_c|}
\sum_{Q_i\in\mathcal Q_c}
\max_{j=1,\ldots,N}
\operatorname{sim}(\mathbf q_i,\mathbf d_j),
& \text{Category-shared}.
\end{cases}
\label{eq:bq}
\end{equation}
Accordingly, we define the \textit{Image-only Retrieval Loss} as
\begin{equation}
\mathcal{L}_{r}
=
\max(0,b^{\star}+\kappa-s_i).
\label{eq:retrieval_loss}
\end{equation}

This objective drives the image similarity above the clean retrieval reference by the prescribed margin, improving the retrieval rankings of poisoned images. 
% This optimization operates only on image pixels, without attacker-controlled text optimization, predefined target answers, or downstream VLM gradients.

\subsection{Caption Preservation Objective}
\label{sec:optimization_objective}

\begin{figure}[t]
    \centering
    \includegraphics[width=\linewidth]{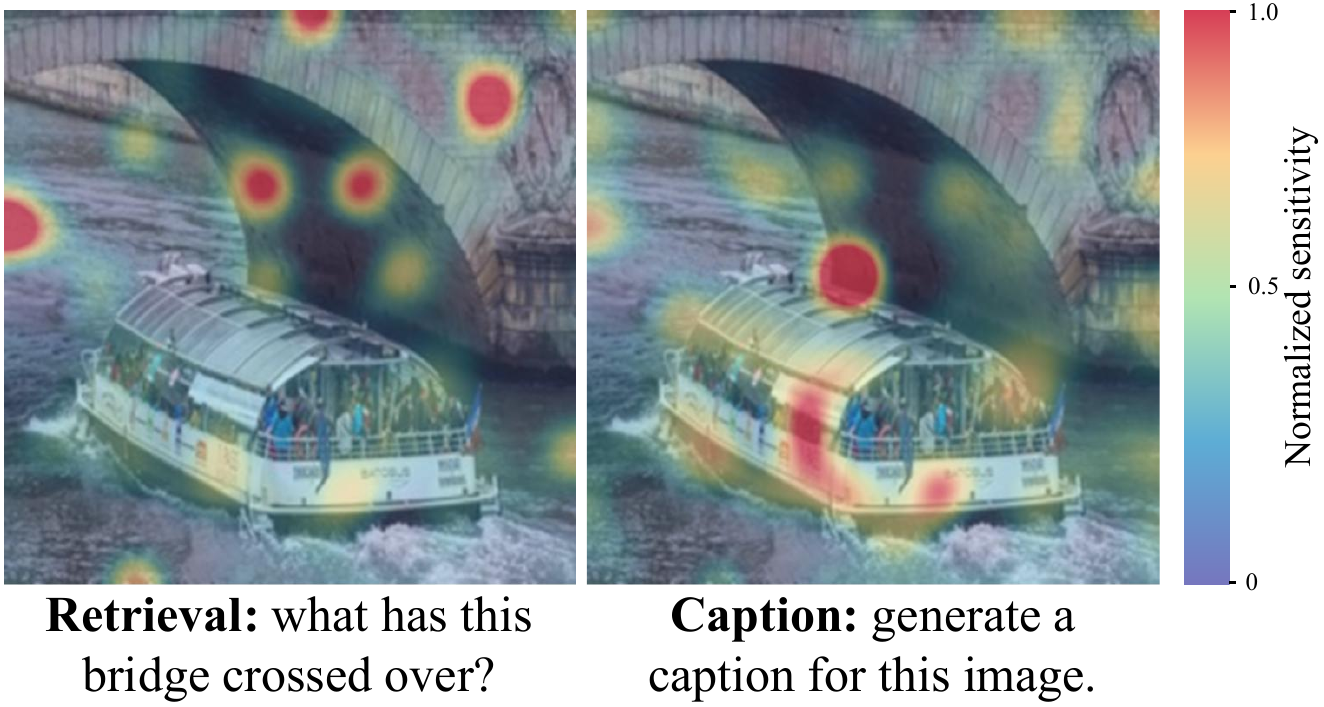}
    \caption{Occlusion-sensitivity maps for retrieval and caption generation, highlighting their distinct sensitive regions.}
    \label{fig:caption}
\end{figure}

During the optimization described above, the perturbation may alter the original visual semantics, causing the generated captions to deviate from the actual image content and making poisoned records conspicuous for humans. Since the attacker cannot control the paired captions, preserving caption generation accuracy is also important during the optimization. As illustrated in Fig.~\ref{fig:caption}, retrieval and caption generation exhibit distinct but partially overlapping visual sensitivity regions. This indicates that the two tasks respond differently to image perturbations. While retrieval-oriented optimization may also affect caption generation, their different sensitivity patterns provide an opportunity to manipulate retrieval representations while preserving the original caption semantics.

Motivated by this observation, we introduce a \textit{Caption Preservation Objective} to preserve the caption fidelity of optimized images. Specifically, we first use the KB's captioning model $\mathcal{C}$ to generate a reference caption $t_i^0=\mathcal{C}(x_i^0)$ for each source image $x_i^0$. We then use this caption as the optimization reference, encouraging the captioning model to retain a high probability of generating the original description when given the poisoned image $x_i$. Accordingly, the \textit{Caption Preservation Loss} is defined as
\begin{equation}
\mathcal{L}_{\mathrm{c}}
=
-\frac{1}{T_i}
\sum_{t=1}^{T_i}
\log p_{\mathcal{C}}
(t_{i,t}^0\mid t_{i,<t}^0,x_i),
\label{eq:caption_loss}
\end{equation}
where $T_i$ denotes the number of tokens in the reference caption, $t_{i,<t}^0$ represents the preceding caption tokens, and $p_{\mathcal{C}}$ denotes the probability of generating the current token given the poisoned image and preceding tokens.

Additionally, to keep the images looking natural, we adopt a \textit{Style Objective} consisting of LPIPS~\cite{zhang2018lpips} and Total Variation (TV)~\cite{rudin1992tv}, defined as $\mathcal{L}_{\mathrm{s}}=\mathcal{L}_{\mathrm{lpips}}+\mathcal{L}_{\mathrm{tv}}$, which penalize perceptual differences and spatial variations in the perturbation, respectively. Finally, we combine these three losses into the overall objective:
\begin{equation}
\mathcal{L}_{\mathrm{total}}
=
\mathcal{L}_{\mathrm{r}}
+\mu\mathcal{L}_{\mathrm{c}}
+\lambda\mathcal{L}_{\mathrm{s}},
\label{eq:total_loss}
\end{equation}
where $\mu$ and $\lambda$ control the weights.

\section{Evaluation}
\label{sec:evaluation}

\subsection{Evaluation Setups}
\label{sec:evaluation_setups}

\begin{table*}[t]
\centering
\small
\setlength{\tabcolsep}{4pt}
\caption{Attack results under equal-weight image-text fusion (\%). Each paired value is reported as Rule/LLM. N-Acc., C-Acc., and P-Acc. denote No-RAG, Clean-RAG, and Poisoned-RAG accuracy, respectively.}
\label{tab:attack_fusion}
\begin{tabular*}{\textwidth}{@{\extracolsep{\fill}}lc*{3}{ccc}@{}}
\toprule
\multirow{2}{*}{VLM} &
\multirow{2}{*}{N-Acc.} &
\multicolumn{3}{c}{UniIR} &
\multicolumn{3}{c}{SigLIP} &
\multicolumn{3}{c}{CLIP} \\
\cmidrule(lr){3-5}\cmidrule(lr){6-8}\cmidrule(lr){9-11}
& & C-Acc. & P-Acc. & ASR
& C-Acc. & P-Acc. & ASR
& C-Acc. & P-Acc. & ASR \\
\midrule

\multicolumn{11}{l}{\textbf{Fixed-query}} \\
\addlinespace[2pt]

LLaVA-1.6 & 8.7/14.6
& 32.4/39.5 & 5.3/14.8 & 93.7/85.9
& 29.9/38.0 & 15.8/25.7 & 71.4/65.9
& 17.7/23.7 & 6.3/16.3 & 81.0/70.6 \\

Qwen2.5-VL & 20.4/28.5
& 34.4/44.3 & 17.3/30.8 & 89.8/80.8
& 25.6/39.2 & 19.4/31.9 & 69.9/64.2
& 23.1/33.4 & 15.5/29.3 & 68.6/64.0 \\

InternVL3.5 & 9.4/18.8
& 33.4/40.9 & 8.7/18.9 & 90.5/82.1
& 27.1/36.6 & 16.6/27.4 & 72.1/66.3
& 17.4/23.3 & 7.0/14.7 & 80.7/83.9 \\

DeepSeek-VL2 & 15.8/22.8
& 29.4/37.8 & 8.3/16.5 & 96.4/91.7
& 24.3/33.4 & 16.3/23.7 & 69.9/67.2
& 18.8/26.3 & 9.8/17.1 & 78.6/77.3 \\

\midrule
\textbf{Average} & \textbf{13.6/21.2}
& \textbf{32.4/40.6} & \textbf{9.9/20.3} & \textbf{92.6/85.1}
& \textbf{26.7/36.8} & \textbf{17.0/27.2} & \textbf{70.8/65.9}
& \textbf{19.3/26.7} & \textbf{9.7/19.4} & \textbf{77.2/74.0} \\

\midrule
\multicolumn{11}{l}{\textbf{Category-shared}} \\
\addlinespace[2pt]

LLaVA-1.6 & 11.4/15.3
& 35.1/39.5 & 27.0/30.3 & 67.3/64.5
& 37.3/45.0 & 35.3/40.5 & 34.7/41.4
& 26.8/29.1 & 23.8/28.5 & 44.0/42.6 \\

Qwen2.5-VL & 19.7/25.7
& 35.8/44.2 & 31.9/38.5 & 61.7/52.4
& 34.5/41.2 & 34.3/41.5 & 42.5/53.5
& 34.2/40.8 & 28.1/36.2 & 50.0/39.4 \\

InternVL3.5 & 13.8/17.0
& 37.6/41.6 & 28.4/33.0 & 74.1/70.8
& 40.0/45.5 & 37.2/42.3 & 34.8/36.2
& 30.8/32.7 & 24.6/27.7 & 44.4/46.2 \\

DeepSeek-VL2 & 18.9/22.4
& 34.6/41.1 & 27.0/32.8 & 75.7/68.8
& 34.5/38.5 & 33.8/39.3 & 38.6/51.6
& 33.2/39.3 & 26.6/30.8 & 54.2/48.6 \\

\midrule
\textbf{Average} & \textbf{16.0/20.1}
& \textbf{35.8/41.6} & \textbf{28.6/33.7} & \textbf{69.7/64.1}
& \textbf{36.6/42.6} & \textbf{35.2/40.9} & \textbf{37.7/45.7}
& \textbf{31.3/35.5} & \textbf{25.8/30.8} & \textbf{48.2/44.2} \\

\bottomrule
\end{tabular*}
\end{table*}

\textbf{Datasets.} We evaluate FAIRP on InfoSeek~\cite{chen2023can} Task 8, whose full validation set contains 17,688 image-question queries and whose KB comprises 481,782 image-text records. For the Fixed-query setting, we use a set of 1,000 queries. For the Category-shared setting, we construct categories from the full validation set using spherical K-Means~\cite{hornik2012spherical} with 128 clusters, 30 iterations, and 3 restarts. After category filtering, we retain 25 categories containing 740 queries in total, with at most 30 queries selected per category. Retrieval in all experiments is performed over the entire KB.

\textbf{Models.} Main experiments employ three retrievers: UniIR CLIP-SF Large~\cite{wei2024uniir}, SigLIP SO400M patch14-384~\cite{zhai2023sigmoid}, and OpenAI CLIP ViT-B/16~\cite{radford2021learning}. We use Qwen2-VL-7B-Instruct~\cite{wang2024qwen2} as the KB's captioning model in main experiments. To further evaluate the Caption Preservation Objective, we additionally use BLIP ViT-B/16~\cite{li2022blip} and CoCa ViT-L/14~\cite{yu2022coca} as retrievers, with each model also serving as its own KB's captioning model. The downstream VLMs include LLaVA-1.6-Mistral-7B~\cite{liu2024llavanext}, Qwen2.5-VL-3B-Instruct~\cite{bai2025qwen25vltechnicalreport}, InternVL3\_5-8B-Instruct~\cite{wang2025internvl3}, and DeepSeek-VL2-Small~\cite{wu2024deepseek}. We use Phi-4-mini-instruct~\cite{abouelenin2025phi} as the LLM judge~\cite{zheng2023judging} to evaluate the downstream VLM's answer correctness. 

\textbf{Retriever settings.} We set the number of retrieved evidence records to $K=3$ and the number of poisoned images to $P=3$ for each query. For category-shared experiments, we inject all the 75 poisoned records from all categories into the KB. We use FAISS~\cite{douze2025faiss} to rank records by the inner product of normalized representations and evaluate three retrieval configurations: image-only retrieval, equal-weight image-text fusion, and Multi-Query retrieval. For Multi-Query Retrieval, we construct four query branches using the original question and three semantically equivalent reformulations, while keeping the query image unchanged. All four branches employ equal-weight image-text fusion. Each branch retrieves the top 20 records, of which the top 10 are used for Reciprocal Rank Fusion~\cite{cormack2009rrf} with a rank constant of 60. And the final top-$K$ records will be returned as the retrieval evidence.

\textbf{Metrics.} We evaluate retrieval performance and downstream question-answering effectiveness using these metrics:
\begin{itemize}
    \item \textit{Ground-truth evidence recall (R@k)} is the percentage of queries for which at least one officially annotated positive record appears among the top-k retrieved records.
    \item \textit{Poison retrieval hit rate (H@k)} is the percentage of queries for which at least one injected poisoned record appears among the top-k retrieved records.
    \item \textit{Answer accuracy (Accuracy)} is reported separately under the No-RAG (N-Acc.), Clean-RAG (C-Acc.), and Poisoned-RAG (P-Acc.) settings, using both rule-based matching and LLM judging.
    \item \textit{Attack success rate (ASR)} is evaluated on queries answered incorrectly under No-RAG but correctly under Clean-RAG. Among these queries, an attack is successful if at least one poisoned record enters the top-$K$ evidence and the Poisoned-RAG answer is incorrect. Rule-based matching and LLM judging independently determine the eligible query sets and the corresponding ASR.
\end{itemize}

\textbf{Implementation details.} We optimize poisoned image pixels using Adam~\cite{adam}. All experiments use a learning rate of 0.005, $\mu = 0.1$, $\lambda=0.1$, $\kappa=0.3$, with 500 optimization steps by default. Pixel values are constrained to $[0,1]$ throughout optimization. All experiments are conducted on an NVIDIA H100.

%%%%%%%%%%%%%%%%%%%%%%%%%%%%%%%%%%%%%%%%%%%%%%%%%%%%%%%%%%%%%%%%%%%%%%
\subsection{Main Results}
\label{sec:main_results}
\begin{table}[t]
\centering
\small
\caption{Caption retention and attack performance without and with $\mathcal L_c$ (\%). Each pair reports results without / with $\mathcal L_c$.}
\label{tab:caption_loss}
\begin{tabular}{lccc}
\toprule
Model & CRR & ASR (Rule) & ASR (LLM) \\
\midrule
BLIP & 1.8 / 73.2 & 49.4 / 43.2 & 54.8 / 52.6 \\
CoCa & 0.4 / 79.4 & 49.3 / 52.0 & 56.3 / 58.1 \\
\bottomrule
\end{tabular}
\end{table}

\textbf{FAIRP can hijack the retriever and induce incorrect VLM responses.} Tab.~\ref{tab:attack_fusion} reports the fixed-query attack results under equal-weight image-text fusion. Clean-RAG outperforms No-RAG across all combinations, increasing average accuracy from 17.4\% to 30.4\%. After poisoning, accuracy decreases across all combinations, with an average reduction of 13.2 pp. The highest ASRs under rule-based matching and LLM judging reach 96.4\% and 91.7\%, respectively, demonstrating that optimizing images alone can undermine the answering benefits provided by clean evidence. Average ASRs under image-only and Multi-Query retrieval are 89.8\% and 60.1\%, respectively; details are provided in supplementary material.

\textbf{Category-shared attacks generalize within the target category.} Tab.~\ref{tab:attack_fusion} reports the category-shared attack results under equal-weight image-text fusion. Average ASRs across the three retrievers range from 41.7\% to 66.9\%, with the highest ASR reaching 75.7\%. Injecting only 75 shared poisoned images across 25 categories still induces incorrect answers for multiple queries, demonstrating that attack coverage can be extended without constructing poisoned images separately for each fixed query. This further indicates that image-only attack exhibits generalization across different queries within the same category. Moreover, under image-only retrieval, average ASRs across the three retrievers range from 28.6\% to 43.2\%; details are provided in supplementary material.

%%%%%%%%%%%%%%%%%%%%%%%%%%%%%%%%%%%%%%%%%%%%%%%%%%%%%%%%%%%%%%%%%%%%%%
\subsection{Caption Preservation Analysis}
\label{sec:caption_evaluation}

\begin{table}[t]
\centering
\footnotesize
\caption{Attack retention rate (\%) after question reformulation. Each entry reports ARR under rule-based matching / LLM judging, averaged over two reformulations.}
\label{tab:question_reformulation}
\begin{tabular*}{\columnwidth}{@{\extracolsep{\fill}}lccc@{}}
\toprule
VLM & UniIR & SigLIP & CLIP \\
\midrule
LLaVA-1.6    & 97.2 / 92.2 & 77.6 / 75.5 & 86.3 / 85.1 \\
Qwen2.5-VL   & 95.9 / 87.4 & 79.7 / 76.2 & 81.3 / 76.3 \\
InternVL3.5  & 96.6 / 89.9 & 78.1 / 69.7 & 84.8 / 81.4 \\
DeepSeek-VL2 & 96.3 / 92.7 & 76.3 / 75.2 & 84.8 / 87.9 \\
\midrule
\textbf{Average}
& \textbf{96.5 / 90.6}
& \textbf{77.9 / 74.2}
& \textbf{84.3 / 82.7} \\
\bottomrule
\end{tabular*}
\end{table}

\textbf{$\mathcal L_c$ is effective when the retriever and captioner share the same model.} We further evaluate BLIP and CoCa, each serving as both the retriever and captioner, with and without $\mathcal L_c$. As shown in Tab.~\ref{tab:caption_loss}, Caption Retention Rate (CRR), measured by rule-based character-level comparison of source and poisoned-image captions, increases from 1.8\% to 73.2\% for BLIP and from 0.4\% to 79.4\% for CoCa. Meanwhile, ASR remains broadly stable across both models, indicating that our attack can manipulate retrieval while substantially preserving caption outputs even when the retriever and captioner share the same model.

As shown in \cref{fig:caption}, we further analyze the visual sensitivity differences between retrieval and caption generation on 1,000 images. The average overlap between the most sensitive regions of the two tasks is only 30.9\%, indicating substantially different but partially overlapping visual regions. This difference may arise because retrieval relies on query-conditioned visual-textual alignment, whereas caption generation focuses on visual cues that support a coherent description of the overall image. Consequently, perturbations that substantially affect retrieval representations do not necessarily have the same impact on generated captions. This partial separation offers an opportunity to manipulate retrieval while largely preserving caption generation (Sec.~\ref{sec:optimization_objective}); accordingly, we constrain changes to caption generation while improving retrieval competitiveness. Vis-Poison~\cite{liang2026vis} demonstrates that captioning and query-conditioned visual tasks rely on different visual cues, and we further highlight the distinct but partially overlapping visual sensitivities between retrieval and caption generation.
\begin{table}[t]
\centering
\footnotesize
\setlength{\tabcolsep}{2.5pt}
\caption{Transfer from InfoSeek to WebQA (\%). Image weight refers to candidate representations; WebQA queries are text-only. ASR is evaluated using the LLM judge.}
\label{tab:cross_dataset_transfer}
\begin{tabular*}{\columnwidth}{@{\extracolsep{\fill}}clccccc@{}}
\toprule
\multirow{2}{*}{\shortstack{Image\\weight (\%)}} &
\multirow{2}{*}{Retriever} &
\multirow{2}{*}{\shortstack{H@3}} &
\multicolumn{4}{c}{ASR} \\
\cmidrule(lr){4-7}
& & & LLaVA & Qwen & Intern & DeepSeek \\
\midrule
\multirow{3}{*}{100}
& UniIR  & 93.7 & 61.1 & 50.3 & 54.0 & 60.0 \\
& SigLIP & 85.3 & 62.0 & 48.6 & 55.1 & 55.9 \\
& CLIP   & 98.5 & 67.3 & 58.2 & 64.9 & 68.5 \\
\midrule
\multirow{3}{*}{75}
& UniIR  & 62.9 & 38.7 & 33.4 & 30.1 & 29.6 \\
& SigLIP & 40.9 & 18.5 & 19.0 & 16.2 & 16.5 \\
& CLIP   & 80.9 & 51.7 & 46.5 & 47.8 & 54.2 \\
\midrule
\multirow{3}{*}{50}
& UniIR  & 18.1 & 3.9 & 4.4 & 4.0 & 4.4 \\
& SigLIP & 17.9 & 1.6 & 2.9 & 3.5 & 3.4 \\
& CLIP   & 27.3 & 13.6 & 8.7 & 7.8 & 8.8 \\
\bottomrule
\end{tabular*}
\end{table}

%%%%%%%%%%%%%%%%%%%%%%%%%%%%%%%%%%%%%%%%%%%%%%%%%%%%%%%%%%%%%%%%%%%%%%
\subsection{Baseline Comparison and Analysis}
\label{sec:baseline}

We adapt two attack families under a unified evaluation framework.
\textbf{Image with malicious text injection.} These methods improve retrieval rankings by directly copying the query image or selecting query-related images, while constructing malicious paired text (e.g., ``Sorry'' or ``I don't know'') to mislead the downstream VLM. However, under our threat model, the attacker can only submit images and cannot control the paired text. The KB automatically generates captions through its normal ingestion process. Consequently, even if the poisoned images successfully enter the top-$K$ evidence, the absence of attacker-crafted malicious text substantially weakens their ability to mislead the VLM, resulting in an ASR of only 2.3\%.

\begin{figure}[t]
\centering
\includegraphics[width=\columnwidth]{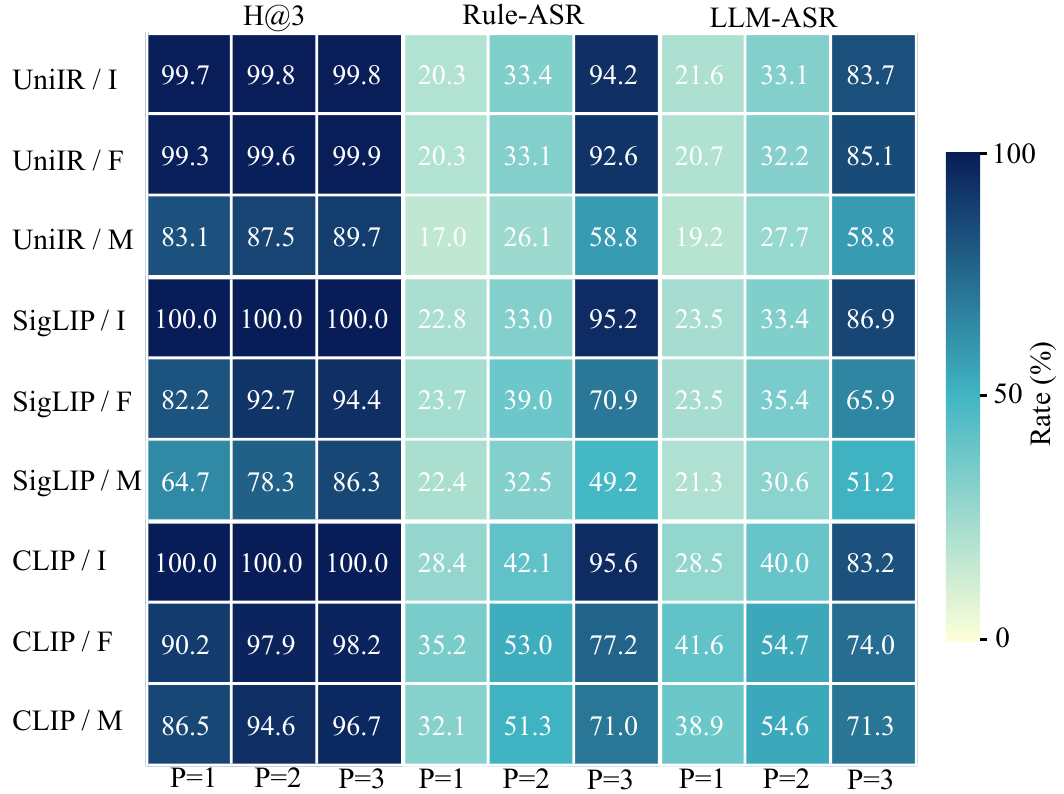}
\caption{Effect of poisoned image count $P$. I, F, and M denote image-only retrieval, equal-weight image-text fusion, and Multi-Query retrieval, respectively. ASRs are averaged over four VLMs.}
\label{fig:poison_quantity}
\end{figure}

\textbf{Semantic image editing.} These methods use an MLLM to generate editing instructions and an image editor to alter visual facts. However, they require query-specific source images, planning, editing, and verification, which cost several times as much as ours. Moreover, 801 of our 1,000 queries concern non-visual facts (e.g., the construction year of a building, the author of an artwork, the city to which a building belongs), which cannot readily be manipulated through visual edits. Importantly, these methods typically tailor each edited image to a specific query and predefined incorrect answer. Even reformulating the question while keeping the query image unchanged generally requires a new semantic edit and attack plan. In contrast, FAIRP can reuse the same poisoned images under question reformulation (Sec.~\ref{sec:question_reformulation}).

\begin{figure*}[t]
\centering
\includegraphics[width=\textwidth]{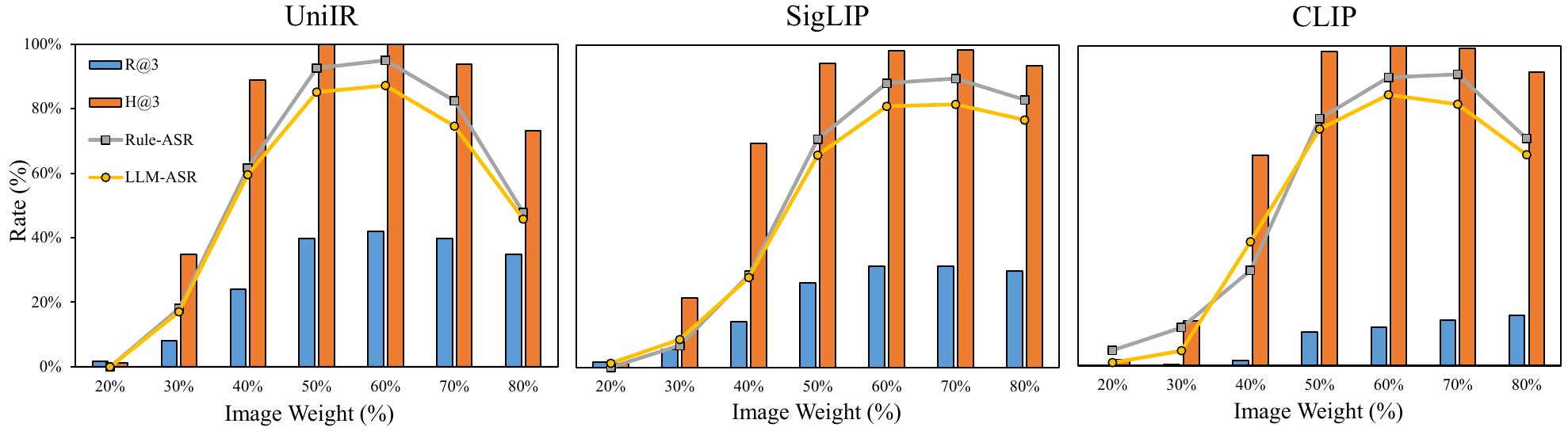}
\caption{Effect of image weight in image-text fusion. R@3 denotes ground-truth evidence recall, and H@3 denotes the poison retrieval hit rate. ASRs are averaged over four VLMs.}
\label{fig:fusion_weight}
\end{figure*}

%%%%%%%%%%%%%%%%%%%%%%%%%%%%%%%%%%%%%%%%%%%%%%%%%%%%%%%%%%%%%%%%%%%%%%
\subsection{Robustness against Question Reformulation}
\label{sec:question_reformulation}

To examine whether FAIRP relies on specific query wording, we evaluate its effectiveness under semantically equivalent question reformulations. Under equal-weight image-text fusion, we select originally successful attacks among eligible queries (answered incorrectly under No-RAG but correctly under Clean-RAG). For each selected question, we generate two semantically equivalent reformulations by changing its wording or word order, while reusing the original poisoned images and paired text without reoptimization. We define attack retention rate (ARR) as the proportion of originally successful cases that remain successful after reformulation. Tab.~\ref{tab:question_reformulation} reports results averaged over the two reformulations, with average ARRs ranging from 74.2\% to 96.5\%, demonstrating that our attack retains substantial effectiveness under changes in question phrasing.

%%%%%%%%%%%%%%%%%%%%%%%%%%%%%%%%%%%%%%%%%%%%%%%%%%%%%%%%%%%%%%%%%%%%%%
\subsection{Transferability}
\label{sec:transferability}

\textbf{Cross-dataset transfer is effective but sensitive to fusion weights.} We transfer the poisoned images, originally constructed on InfoSeek under equal-weight image-text fusion, to all queries in WebQA Task 2~\cite{chang2022webqa}, without reoptimizing the images. Tab.~\ref{tab:cross_dataset_transfer} shows that using image-only candidate representations at the target achieves an H@3 of 85.3\%--98.5\% and ASRs of 48.6\%--68.5\% across the four VLMs. When the target KB uses image/text weights of 75\% / 25\%, ASRs range from 16.2\% to 54.2\%, but decrease to 1.6\%-13.6\% under equal-weight fusion. These results demonstrate cross-dataset transfer while revealing its strong dependence on the relative contributions of image and text features in target candidate representations.

\textbf{Cross-retriever transfer remains challenging.} Under equal-weight image-text fusion, cross-retriever transfer achieves an H@3 of 3\%. This limited transferability likely stems from differences in feature representations, embedding geometries, and similarity distributions across retrievers, which make the optimization landscape highly model-dependent. In particular, perturbations that effectively improve image-query alignment in the source retriever may induce substantially different representation shifts in another one, failing to maintain their retrieval advantage.

%%%%%%%%%%%%%%%%%%%%%%%%%%%%%%%%%%%%%%%%%%%%%%%%%%%%%%%%%%%%%%%%%%%%%%
\subsection{Ablation Study}
\label{sec:ablation}

\textbf{Number of poisoned images.} We vary the number of injected poisoned images $P\in\{1,2,3\}$, reusing existing images and their paired captions without reoptimization. Fig.~\ref{fig:poison_quantity} shows that ASR increases with $P$ across all configurations. In particular, under SigLIP image-only retrieval, H@3 remains at 100\%, while rule-based ASR increases from 22.8\% at $P=1$ to 95.2\% at $P=3$. This indicates that retrieving at least one poisoned record does not necessarily suffice to mislead the VLM. Injecting more poisoned records allows them to occupy more top-$K$ evidence positions, potentially reducing the support provided by the remaining clean evidence and thereby increasing attack success.

\textbf{Image-text fusion weights.} We reuse poisoned records constructed under equal-weight fusion and vary only the image weight during retrieval, without reoptimization. As shown in Fig.~\ref{fig:fusion_weight}, ASRs peak at image weights of 60\%-70\% across the three retrievers, with the highest rule-based ASR reaching 95.0\%, before declining at an image weight of 80\%. This non-monotonic trend indicates that increasing the image contribution does not necessarily improve attack effectiveness, which depends on whether the optimized images retain their ranking advantage in the resulting fused representations. Meanwhile, lower image weights reduce H@3 and ASR but also substantially degrade R@3, revealing a trade-off between suppressing attacks and maintaining ground-truth evidence recall.

%%%%%%%%%%%%%%%%%%%%%%%%%%%%%%%%%%%%%%%%%%%%%%%%%%%%%%%%%%%%%%%%%%%%%%
% \subsection{Case Study}
% \label{sec:case_study}

% Fig.~\ref{fig:case_study} illustrates how poisoned records displace useful evidence. Under No-RAG, the VLM responds with ``Unknown''; after Clean-RAG retrieves grape records containing \textit{Vitis labrusca}, the model answers correctly. After poisoning, three irrelevant records occupy the top-3 results, causing the model to return ``Unknown'' again. These records and their automatically generated captions do not provide false knowledge tailored to the question. Instead, they undermine retrieval augmentation by displacing evidence supporting the correct answer. This case demonstrates that the attack can turn a correct response into an answering failure without specifying an incorrect target answer.

% % \input{figure/fig-case-study}

\section{Discussion}

\textbf{Implications for multimodal RAG security.} Our findings show that image-caption consistency alone is insufficient to ensure reliable retrieval. By submitting images alone, an attacker can manipulate retrieval rankings and introduce irrelevant records into the evidence context, thereby undermining the effectiveness of RAG. Therefore, multimodal RAG security should consider not only image-caption consistency but also the relevance of retrieved evidence to user queries. This highlights the need for query-conditioned evidence verification and more robust retrieval mechanisms.
% This motivates query-conditioned evidence verification and more robust retrieval mechanisms.

\textbf{Limitations.} Our attack assumes white-box access to the target retriever. Moreover, our evaluation focuses primarily on InfoSeek and a selected set of retrievers and VLMs. Future work could explore black-box attack settings and broader multimodal RAG benchmarks.
% The caption retention rate is measured using rule-based character-level comparison, which captures textual retention but does not fully characterize semantic fidelity. 
\section{Conclusion}

In this work, we propose FAIRP, an image-only poisoning framework for multimodal RAG that manipulates evidence selection through pixel-level optimization without controlling paired text, predefining target answers, or accessing the downstream VLM. Experiments validate fixed-query and category-shared attacks and the robustness. Building on the distinct but partially overlapping visual sensitivities of retrieval and caption generation, we introduce a caption-preservation objective that improves caption retention in shared retriever-captioner settings while maintaining attack effectiveness. This novel attack exposes a security risk in multimodal RAG.

\section*{LLM Usage Statement}

In this work, LLM tools (ChatGPT) were used only for language polishing, LaTeX formatting, and coding assistance. All substantive research decisions and intellectual contributions were made by the authors. All LLM-assisted outputs were reviewed by the authors, who take full responsibility for the final work.

% \clearpage
{
    \small
    \bibliographystyle{ieeenat_fullname}
    \bibliography{main}

\begin{thebibliography}{49}
\providecommand{\natexlab}[1]{#1}
\providecommand{\url}[1]{\texttt{#1}}
\expandafter\ifx\csname urlstyle\endcsname\relax
  \providecommand{\doi}[1]{doi: #1}\else
  \providecommand{\doi}{doi: \begingroup \urlstyle{rm}\Url}\fi

\bibitem[Abouelenin et~al.(2025)Abouelenin, Ashfaq, Atkinson, Awadalla, Bach, Bao, Benhaim, Cai, Chaudhary, Chen, et~al.]{abouelenin2025phi}
Abdelrahman Abouelenin, Atabak Ashfaq, Adam Atkinson, Hany Awadalla, Nguyen Bach, Jianmin Bao, Alon Benhaim, Martin Cai, Vishrav Chaudhary, Congcong Chen, et~al.
\newblock Phi-4-mini technical report: Compact yet powerful multimodal language models via mixture-of-loras.
\newblock \emph{arXiv preprint arXiv:2503.01743}, 2025.

\bibitem[Bai et~al.(2025)Bai, Chen, Liu, Wang, Ge, Song, Dang, Wang, Wang, Tang, Zhong, Zhu, Yang, Li, Wan, Wang, Ding, Fu, Xu, Ye, Zhang, Xie, Cheng, Zhang, Yang, Xu, and Lin]{bai2025qwen25vltechnicalreport}
Shuai Bai, Keqin Chen, Xuejing Liu, Jialin Wang, Wenbin Ge, Sibo Song, Kai Dang, Peng Wang, Shijie Wang, Jun Tang, Humen Zhong, Yuanzhi Zhu, Mingkun Yang, Zhaohai Li, Jianqiang Wan, Pengfei Wang, Wei Ding, Zheren Fu, Yiheng Xu, Jiabo Ye, Xi Zhang, Tianbao Xie, Zesen Cheng, Hang Zhang, Zhibo Yang, Haiyang Xu, and Junyang Lin.
\newblock Qwen2.5-vl technical report, 2025.

\bibitem[Bailey et~al.(2024)Bailey, Ong, Russell, and Emmons]{bailey2024runtime}
Luke Bailey, Euan Ong, Stuart Russell, and Scott Emmons.
\newblock Image hijacks: adversarial images can control generative models at runtime.
\newblock In \emph{Proceedings of the 41st International Conference on Machine Learning}, 2024.

\bibitem[Chang et~al.(2022)Chang, Cao, Narang, Gao, Suzuki, and Bisk]{chang2022webqa}
Yingshan Chang, Guihong Cao, Mridu Narang, Jianfeng Gao, Hisami Suzuki, and Yonatan Bisk.
\newblock Web{QA}: Multihop and multimodal {QA}.
\newblock In \emph{2022 IEEE/CVF Conference on Computer Vision and Pattern Recognition (CVPR)}, pages 16474--16483, 2022.

\bibitem[Chen et~al.(2022)Chen, Hu, Chen, Verga, and Cohen]{chen2022murag}
Wenhu Chen, Hexiang Hu, Xi Chen, Pat Verga, and William Cohen.
\newblock Murag: Multimodal retrieval-augmented generator for open question answering over images and text.
\newblock In \emph{Proceedings of the 2022 Conference on Empirical Methods in Natural Language Processing}, pages 5558--5570, 2022.

\bibitem[Chen et~al.(2023)Chen, Hu, Luan, Sun, Changpinyo, Ritter, and Chang]{chen2023can}
Yang Chen, Hexiang Hu, Yi Luan, Haitian Sun, Soravit Changpinyo, Alan Ritter, and Ming-Wei Chang.
\newblock Can pre-trained vision and language models answer visual information-seeking questions?
\newblock In \emph{Proceedings of the 2023 Conference on Empirical Methods in Natural Language Processing}, pages 14948--14968, 2023.

\bibitem[Cormack et~al.(2009)Cormack, Clarke, and Buettcher]{cormack2009rrf}
Gordon~V. Cormack, Charles L~A Clarke, and Stefan Buettcher.
\newblock Reciprocal rank fusion outperforms condorcet and individual rank learning methods.
\newblock In \emph{Proceedings of the 32nd International ACM SIGIR Conference on Research and Development in Information Retrieval}, page 758–759, New York, NY, USA, 2009. Association for Computing Machinery.

\bibitem[Douze et~al.(2026)Douze, Guzhva, Deng, Johnson, Szilvasy, Mazar{\'e}, Lomeli, Hosseini, and J{\'e}gou]{douze2025faiss}
Matthijs Douze, Alexandr Guzhva, Chengqi Deng, Jeff Johnson, Gergely Szilvasy, Pierre-Emmanuel Mazar{\'e}, Maria Lomeli, Lucas Hosseini, and Herv{\'e} J{\'e}gou.
\newblock The faiss library.
\newblock \emph{IEEE Transactions on Big Data}, 2026.

\bibitem[Faysse et~al.(2025)Faysse, Sibille, Wu, Omrani, Viaud, Hudelot, and Colombo]{faysse2025colpali}
Manuel Faysse, Hugues Sibille, Tony Wu, Bilel Omrani, Gautier Viaud, C{\'e}line Hudelot, and Pierre Colombo.
\newblock Colpali: Efficient document retrieval with vision language models.
\newblock In \emph{International Conference on Learning Representations}, 2025.

\bibitem[Gong et~al.(2025)Gong, Chen, Liu, Chen, Yu, Lu, Wang, and Liu]{gong2025topic}
Yuyang Gong, Zhuo Chen, Jiawei Liu, Miaokun Chen, Fengchang Yu, Wei Lu, XiaoFeng Wang, and Xiaozhong Liu.
\newblock Topic-{F}lip{RAG}: Topic-orientated adversarial opinion manipulation attacks to retrieval-augmented generation models.
\newblock In \emph{34th USENIX Security Symposium (USENIX Security 25)}, pages 3807--3826, 2025.

\bibitem[Greshake et~al.(2023)Greshake, Abdelnabi, Mishra, Endres, Holz, and Fritz]{greshake2023not}
Kai Greshake, Sahar Abdelnabi, Shailesh Mishra, Christoph Endres, Thorsten Holz, and Mario Fritz.
\newblock Not what you've signed up for: Compromising real-world llm-integrated applications with indirect prompt injection.
\newblock In \emph{Proceedings of the 16th ACM workshop on artificial intelligence and security}, pages 79--90, 2023.

\bibitem[Guu et~al.(2020)Guu, Lee, Tung, Pasupat, and Chang]{guu2020retrieval}
Kelvin Guu, Kenton Lee, Zora Tung, Panupong Pasupat, and Mingwei Chang.
\newblock Retrieval augmented language model pre-training.
\newblock In \emph{International conference on machine learning}, pages 3929--3938. PMLR, 2020.

\bibitem[Ha et~al.(2026)Ha, Zhan, Kim, Bralios, Sanniboina, Peng, Chang, Kang, and Ji]{ha-etal-2026-mm}
Hyeonjeong Ha, Qiusi Zhan, Jeonghwan Kim, Dimitrios Bralios, Saikrishna Sanniboina, Nanyun Peng, Kai-Wei Chang, Daniel Kang, and Heng Ji.
\newblock {MM}-{P}oison{RAG}: Disrupting multimodal {RAG} with local and global knowledge poisoning attacks.
\newblock In \emph{Proceedings of the 64th Annual Meeting of the {A}ssociation for {C}omputational {L}inguistics (Volume 1: Long Papers)}, pages 33804--33826, 2026.

\bibitem[Hornik et~al.(2012)Hornik, Feinerer, Kober, and Buchta]{hornik2012spherical}
Kurt Hornik, Ingo Feinerer, Martin Kober, and Christian Buchta.
\newblock Spherical k-means clustering.
\newblock \emph{Journal of statistical software}, 50\penalty0 (10):\penalty0 1--22, 2012.

\bibitem[Hu et~al.(2025)Hu, Gu, Dou, Fayyaz, Lu, Chang, and Peng]{hu2025mrag}
Wenbo Hu, Jia-Chen Gu, Zi-Yi Dou, Mohsen Fayyaz, Pan Lu, Kai-Wei Chang, and Nanyun~Violet Peng.
\newblock Mrag-bench: Vision-centric evaluation for retrieval-augmented multimodal models.
\newblock In \emph{International Conference on Learning Representations}, 2025.

\bibitem[Hu et~al.(2023)Hu, Iscen, Sun, Wang, Chang, Sun, Schmid, Ross, and Fathi]{hu2023reveal}
Ziniu Hu, Ahmet Iscen, Chen Sun, Zirui Wang, Kai-Wei Chang, Yizhou Sun, Cordelia Schmid, David~A Ross, and Alireza Fathi.
\newblock Reveal: Retrieval-augmented visual-language pre-training with multi-source multimodal knowledge memory.
\newblock In \emph{2023 IEEE/CVF Conference on Computer Vision and Pattern Recognition (CVPR)}, pages 23369--23379. IEEE, 2023.

\bibitem[Izacard and Grave(2021)]{izacard2021leveraging}
Gautier Izacard and Edouard Grave.
\newblock Leveraging passage retrieval with generative models for open domain question answering.
\newblock In \emph{Proceedings of the 16th conference of the european chapter of the association for computational linguistics: main volume}, pages 874--880, 2021.

\bibitem[Karpukhin et~al.(2020)Karpukhin, Oguz, Min, Lewis, Wu, Edunov, Chen, and Yih]{karpukhin2020dense}
Vladimir Karpukhin, Barlas Oguz, Sewon Min, Patrick Lewis, Ledell Wu, Sergey Edunov, Danqi Chen, and Wen-tau Yih.
\newblock Dense passage retrieval for open-domain question answering.
\newblock In \emph{Proceedings of the 2020 conference on empirical methods in natural language processing (EMNLP)}, pages 6769--6781, 2020.

\bibitem[Kingma and Ba(2015)]{adam}
Diederik Kingma and Jimmy Ba.
\newblock Adam: A method for stochastic optimization.
\newblock \emph{International Conference on Learning Representations}, 2015.

\bibitem[Lewis et~al.(2020)Lewis, Perez, Piktus, Petroni, Karpukhin, Goyal, K{\"u}ttler, Lewis, Yih, Rockt{\"a}schel, et~al.]{lewis2020retrieval}
Patrick Lewis, Ethan Perez, Aleksandra Piktus, Fabio Petroni, Vladimir Karpukhin, Naman Goyal, Heinrich K{\"u}ttler, Mike Lewis, Wen-tau Yih, Tim Rockt{\"a}schel, et~al.
\newblock Retrieval-augmented generation for knowledge-intensive nlp tasks.
\newblock \emph{Advances in neural information processing systems}, 33:\penalty0 9459--9474, 2020.

\bibitem[Li et~al.(2022)Li, Li, Xiong, and Hoi]{li2022blip}
Junnan Li, Dongxu Li, Caiming Xiong, and Steven Hoi.
\newblock Blip: Bootstrapping language-image pre-training for unified vision-language understanding and generation.
\newblock In \emph{International conference on machine learning}, pages 12888--12900. PMLR, 2022.

\bibitem[Li et~al.(2023)Li, Li, Savarese, and Hoi]{li2023blip}
Junnan Li, Dongxu Li, Silvio Savarese, and Steven Hoi.
\newblock Blip-2: Bootstrapping language-image pre-training with frozen image encoders and large language models.
\newblock In \emph{International conference on machine learning}, pages 19730--19742. PMLR, 2023.

\bibitem[Liang et~al.(2026)Liang, Chen, Lei, and Miao]{liang2026vis}
Rujin Liang, Zhongpu Chen, Yuhao Lei, and Xin Miao.
\newblock Vis-poison: Poisoning visual knowledge in multimodal retrieval-augmented generation.
\newblock In \emph{Findings of the Association for Computational Linguistics: EMNLP 2026}, pages 7280--7294, 2026.

\bibitem[Liu et~al.(2023)Liu, Li, Wu, and Lee]{liu2023visual}
Haotian Liu, Chunyuan Li, Qingyang Wu, and Yong~Jae Lee.
\newblock Visual instruction tuning.
\newblock \emph{Advances in neural information processing systems}, 36:\penalty0 34892--34916, 2023.

\bibitem[Liu et~al.(2024)Liu, Li, Li, Li, Zhang, Shen, and Lee]{liu2024llavanext}
Haotian Liu, Chunyuan Li, Yuheng Li, Bo Li, Yuanhan Zhang, Sheng Shen, and Yong~Jae Lee.
\newblock Llava-next: Improved reasoning, ocr, and world knowledge, 2024.

\bibitem[Liu et~al.(2026)Liu, Zhou, Dong, Li, and Yu]{liu2026warp}
Hui Liu, Yibo Zhou, Liguo Dong, Weidong Li, and Shui Yu.
\newblock Warp: A word-level backdoor attack targeting rag systems via retrieval corpus poisoning.
\newblock In \emph{Proceedings of the 32nd ACM SIGKDD Conference on Knowledge Discovery and Data Mining V. 1}, pages 867--878, 2026.

\bibitem[Liu et~al.(2025)Liu, Yuan, Tie, Shi, Zhou, Sun, and Gong]{liu2025poisoned}
Yinuo Liu, Zenghui Yuan, Guiyao Tie, Jiawen Shi, Pan Zhou, Lichao Sun, and Neil~Zhenqiang Gong.
\newblock Poisoned-mrag: Knowledge poisoning attacks to multimodal retrieval augmented generation.
\newblock \emph{arXiv preprint arXiv:2503.06254}, 2025.

\bibitem[Moosavi-Dezfooli et~al.(2017)Moosavi-Dezfooli, Fawzi, Fawzi, and Frossard]{moosavi2017universal}
Seyed-Mohsen Moosavi-Dezfooli, Alhussein Fawzi, Omar Fawzi, and Pascal Frossard.
\newblock Universal adversarial perturbations.
\newblock In \emph{2017 IEEE conference on computer vision and pattern recognition (CVPR)}, pages 86--94. IEEE, 2017.

\bibitem[Qi et~al.(2024)Qi, Huang, Panda, Henderson, Wang, and Mittal]{qi2024visual}
Xiangyu Qi, Kaixuan Huang, Ashwinee Panda, Peter Henderson, Mengdi Wang, and Prateek Mittal.
\newblock Visual adversarial examples jailbreak aligned large language models.
\newblock In \emph{Proceedings of the AAAI conference on artificial intelligence}, pages 21527--21536, 2024.

\bibitem[Radford et~al.(2021)Radford, Kim, Hallacy, Ramesh, Goh, Agarwal, Sastry, Askell, Mishkin, Clark, et~al.]{radford2021learning}
Alec Radford, Jong~Wook Kim, Chris Hallacy, Aditya Ramesh, Gabriel Goh, Sandhini Agarwal, Girish Sastry, Amanda Askell, Pamela Mishkin, Jack Clark, et~al.
\newblock Learning transferable visual models from natural language supervision.
\newblock In \emph{International conference on machine learning}, pages 8748--8763. PMLR, 2021.

\bibitem[Rudin et~al.(1992)Rudin, Osher, and Fatemi]{rudin1992tv}
Leonid~I Rudin, Stanley Osher, and Emad Fatemi.
\newblock Nonlinear total variation based noise removal algorithms.
\newblock \emph{Physica D: nonlinear phenomena}, 60\penalty0 (1-4):\penalty0 259--268, 1992.

\bibitem[Shafran et~al.(2025)Shafran, Schuster, and Shmatikov]{shafran2025machine}
Avital Shafran, Roei Schuster, and Vitaly Shmatikov.
\newblock Machine against the {RAG}: Jamming {Retrieval-Augmented} generation with blocker documents.
\newblock In \emph{34th USENIX Security Symposium (USENIX Security 25)}, pages 3787--3806, 2025.

\bibitem[Shereen et~al.(2026)Shereen, Ristea, McFadden, Hasircioglu, Mavroudis, and Hicks]{shereen2026one}
Ezzeldin Shereen, Dan Ristea, Shae McFadden, Burak Hasircioglu, Vasilios Mavroudis, and Chris Hicks.
\newblock One pic is all it takes: Poisoning visual document retrieval augmented generation with a single image.
\newblock \emph{Transactions on Machine Learning Research}, 2026.

\bibitem[Talmor et~al.(2021)Talmor, Yoran, Catav, Lahav, Wang, Asai, Ilharco, Hajishirzi, and Berant]{talmor2021multimodalqa}
Alon Talmor, Ori Yoran, Amnon Catav, Dan Lahav, Yizhong Wang, Akari Asai, Gabriel Ilharco, Hannaneh Hajishirzi, and Jonathan Berant.
\newblock Multimodal{QA}: complex question answering over text, tables and images.
\newblock In \emph{International Conference on Learning Representations}, 2021.

\bibitem[Wallace et~al.(2019)Wallace, Feng, Kandpal, Gardner, and Singh]{wallace2019universal}
Eric Wallace, Shi Feng, Nikhil Kandpal, Matt Gardner, and Sameer Singh.
\newblock Universal adversarial triggers for attacking and analyzing {NLP}.
\newblock In \emph{Proceedings of the 2019 conference on empirical methods in natural language processing and the 9th international joint conference on natural language processing (EMNLP-IJCNLP)}, pages 2153--2162, 2019.

\bibitem[Wang et~al.(2024)Wang, Bai, Tan, Wang, Fan, Bai, Chen, Liu, Wang, Ge, et~al.]{wang2024qwen2}
Peng Wang, Shuai Bai, Sinan Tan, Shijie Wang, Zhihao Fan, Jinze Bai, Keqin Chen, Xuejing Liu, Jialin Wang, Wenbin Ge, et~al.
\newblock Qwen2-vl: Enhancing vision-language model's perception of the world at any resolution.
\newblock \emph{arXiv preprint arXiv:2409.12191}, 2024.

\bibitem[Wang et~al.(2025)Wang, Gao, Gu, Pu, Cui, Wei, Liu, Jing, Ye, Shao, et~al.]{wang2025internvl3}
Weiyun Wang, Zhangwei Gao, Lixin Gu, Hengjun Pu, Long Cui, Xingguang Wei, Zhaoyang Liu, Linglin Jing, Shenglong Ye, Jie Shao, et~al.
\newblock Internvl3.5: Advancing open-source multimodal models in versatility, reasoning, and efficiency.
\newblock \emph{arXiv preprint arXiv:2508.18265}, 2025.

\bibitem[Wei et~al.(2024)Wei, Chen, Chen, Hu, Zhang, Fu, Ritter, and Chen]{wei2024uniir}
Cong Wei, Yang Chen, Haonan Chen, Hexiang Hu, Ge Zhang, Jie Fu, Alan Ritter, and Wenhu Chen.
\newblock Uniir: Training and benchmarking universal multimodal information retrievers.
\newblock In \emph{European Conference on Computer Vision}, pages 387--404. Springer, 2024.

\bibitem[Wu et~al.(2024)Wu, Chen, Pan, Liu, Liu, Dai, Gao, Ma, Wu, Wang, et~al.]{wu2024deepseek}
Zhiyu Wu, Xiaokang Chen, Zizheng Pan, Xingchao Liu, Wen Liu, Damai Dai, Huazuo Gao, Yiyang Ma, Chengyue Wu, Bingxuan Wang, et~al.
\newblock Deepseek-vl2: Mixture-of-experts vision-language models for advanced multimodal understanding.
\newblock \emph{arXiv preprint arXiv:2412.10302}, 2024.

\bibitem[Yang et~al.(2026)Yang, Zheng, Ju, Wang, Ni, Liu, Wang, Huang, and Qi]{yang2026knowledge}
Peiru Yang, Haoran Zheng, Tong Ju, Shiting Wang, Wanchun Ni, Jiajun Liu, Shangguang Wang, Yongfeng Huang, and Tao Qi.
\newblock Knowledge poisoning attacks on medical multi-modal retrieval-augmented generation.
\newblock In \emph{Proceedings of the 64th Annual Meeting of the Association for Computational Linguistics (Volume 1: Long Papers)}, pages 19494--19513, 2026.

\bibitem[Yu et~al.(2022)Yu, Wang, Vasudevan, Yeung, Seyedhosseini, and Wu]{yu2022coca}
Jiahui Yu, Zirui Wang, Vijay Vasudevan, Legg Yeung, Mojtaba Seyedhosseini, and Yonghui Wu.
\newblock Coca: Contrastive captioners are image-text foundation models.
\newblock \emph{arXiv preprint arXiv:2205.01917}, 2022.

\bibitem[Yu et~al.(2025)Yu, Tang, Xu, Cui, Ran, Yan, Liu, Wang, Han, Liu, et~al.]{yu2025visrag}
Shi Yu, Chaoyue Tang, Bokai Xu, Junbo Cui, Junhao Ran, Yukun Yan, Zhenghao Liu, Shuo Wang, Xu Han, Zhiyuan Liu, et~al.
\newblock Visrag: Vision-based retrieval-augmented generation on multi-modality documents.
\newblock In \emph{International Conference on Learning Representations}, 2025.

\bibitem[Zhai et~al.(2023)Zhai, Mustafa, Kolesnikov, and Beyer]{zhai2023sigmoid}
Xiaohua Zhai, Basil Mustafa, Alexander Kolesnikov, and Lucas Beyer.
\newblock Sigmoid loss for language image pre-training.
\newblock In \emph{2023 IEEE/CVF International Conference on Computer Vision (ICCV)}, pages 11941--11952. IEEE, 2023.

\bibitem[Zhan et~al.(2024)Zhan, Liang, Ying, and Kang]{zhan2024injecagent}
Qiusi Zhan, Zhixiang Liang, Zifan Ying, and Daniel Kang.
\newblock Injecagent: Benchmarking indirect prompt injections in tool-integrated large language model agents.
\newblock In \emph{Findings of the Association for Computational Linguistics: ACL 2024}, pages 10471--10506, 2024.

\bibitem[Zhang et~al.(2025)Zhang, Zhang, Lou, Wu, Wang, and Chen]{zhang2025poisonedeye}
Chenyang Zhang, Xiaoyu Zhang, Jian Lou, Kai Wu, Zilong Wang, and Xiaofeng Chen.
\newblock Poisonedeye: Knowledge poisoning attack on retrieval-augmented generation based large vision-language models.
\newblock In \emph{Forty-second International Conference on Machine Learning}, 2025.

\bibitem[Zhang et~al.(2018)Zhang, Isola, Efros, Shechtman, and Wang]{zhang2018lpips}
Richard Zhang, Phillip Isola, Alexei~A Efros, Eli Shechtman, and Oliver Wang.
\newblock The unreasonable effectiveness of deep features as a perceptual metric.
\newblock In \emph{2018 IEEE/CVF conference on computer vision and pattern recognition}, pages 586--595. IEEE, 2018.

\bibitem[Zheng et~al.(2023)Zheng, Chiang, Sheng, Zhuang, Wu, Zhuang, Lin, Li, Li, Xing, et~al.]{zheng2023judging}
Lianmin Zheng, Wei-Lin Chiang, Ying Sheng, Siyuan Zhuang, Zhanghao Wu, Yonghao Zhuang, Zi Lin, Zhuohan Li, Dacheng Li, Eric Xing, et~al.
\newblock Judging llm-as-a-judge with mt-bench and chatbot arena.
\newblock \emph{Advances in neural information processing systems}, 36:\penalty0 46595--46623, 2023.

\bibitem[Zhong et~al.(2023)Zhong, Huang, Wettig, and Chen]{zhong2023poisoning}
Zexuan Zhong, Ziqing Huang, Alexander Wettig, and Danqi Chen.
\newblock Poisoning retrieval corpora by injecting adversarial passages.
\newblock In \emph{Proceedings of the 2023 Conference on Empirical Methods in Natural Language Processing}, pages 13764--13775, 2023.

\bibitem[Zou et~al.(2025)Zou, Geng, Wang, and Jia]{zou2025poisonedrag}
Wei Zou, Runpeng Geng, Binghui Wang, and Jinyuan Jia.
\newblock Poisoned{RAG}: Knowledge corruption attacks to retrieval-augmented generation of large language models.
\newblock In \emph{34th USENIX Security Symposium (USENIX Security 25)}, pages 3827--3844, 2025.

\end{thebibliography}
}

\clearpage
\setcounter{page}{1}
\maketitlesupplementary

\setcounter{table}{0}
\renewcommand{\thetable}{S\arabic{table}}
\setcounter{figure}{0}
\renewcommand{\thefigure}{S\arabic{figure}}

\section{Additional Experiments and Analysis}
\label{appsec:additonal_exp}

\subsection{Additional Main Results}
\label{appsec:additional_main}

\begin{table*}[t]
    \centering
    \caption{Numbers of eligible queries and successful attacks in the main experiments. Each entry is reported as $S/E$, where $S$ denotes the number of successful attacks and $E$ denotes the number of eligible queries. Rule-based Matching and LLM Judging are reported separately.}
    \label{apptab:query}
    \setlength{\tabcolsep}{10pt}
    \renewcommand{\arraystretch}{1.15}
    \begin{tabular}{lcccccc}
        \toprule
        & \multicolumn{2}{c}{UniIR}
        & \multicolumn{2}{c}{SigLIP}
        & \multicolumn{2}{c}{CLIP} \\
        \cmidrule(lr){2-3}
        \cmidrule(lr){4-5}
        \cmidrule(lr){6-7}
        VLM & Rule & LLM & Rule & LLM & Rule & LLM \\
        \midrule
        \multicolumn{7}{l}{\textbf{Fixed-query}} \\
        \midrule
        LLaVA-1.6
        & 236/252 & 225/262 & 157/220 & 162/246 & 81/100 & 72/102 \\
        Qwen2.5-VL
        & 132/147 & 135/167 & 51/73 & 77/120 & 24/35 & 32/50 \\
        InternVL3.5
        & 228/252 & 188/229 & 137/190 & 124/187 & 71/88 & 47/56 \\
        DeepSeek-VL2
        & 134/139 & 144/157 & 65/93 & 78/116 & 33/42 & 34/44 \\
        \midrule
        \multicolumn{7}{l}{\textbf{Category-shared}} \\
        \midrule
        LLaVA-1.6
        & 132/196 & 118/183 & 67/193 & 91/220 & 51/116 & 46/108 \\
        Qwen2.5-VL
        & 74/120 & 75/143 & 48/113 & 68/127 & 54/108 & 50/127 \\
        InternVL3.5
        & 137/185 & 131/185 & 69/198 & 77/213 & 56/126 & 54/117 \\
        DeepSeek-VL2
        & 103/136 & 95/138 & 49/127 & 63/122 & 58/107 & 67/138 \\
        \bottomrule
    \end{tabular}
\end{table*}

\textbf{Eligible Query in the Main Experiments.}
\cref{tab:attack_fusion} in the main paper presents the attack results of FAIRP in different combinations of Retriever and VLM, where the ASR is calculated only on eligible queries. As defined in \cref{sec:evaluation_setups}, an eligible query is one that is answered incorrectly by No-RAG but correctly by Clean-RAG. Among these queries, an attack is considered successful if at least one poisoned record appears in the retrieved Top-\(K\) evidence and the Poisoned-RAG answer is incorrect. Rule-based Matching and LLM Judging independently determine the eligible query sets and the corresponding numbers of successful attacks. To further clarify the actual evaluation counts underlying ASR for each experimental configuration, \cref{apptab:query} summarizes the numbers of eligible queries.

\textbf{Fixed-query attack under image-only retrieval.}
The results in \cref{apptab:fixed_query_image} show that FAIRP remains highly effective when retrieval relies exclusively on visual features, achieving average ASRs of 88.9\%, 91.0\%, and 89.4\% for UniIR, SigLIP, and CLIP, respectively. Notably, this effectiveness is consistent across all four VLMs, with Poisoned-RAG accuracy substantially lower than Clean-RAG accuracy in every model-retriever combination. Since no textual features participate in candidate retrieval, these results demonstrate that manipulating visual representations alone is sufficient to displace relevant evidence, without relying on image-text fusion or attacker-controlled text.

\textbf{Fixed-query attack under Multi-Query retrieval.}
Unlike image-only retrieval, Multi-Query retrieval aggregates signals from multiple query branches, providing more diverse retrieval representations. As shown in \cref{apptab:fixed_query_multiquery}, this setting reduces the average ASR to 58.8\%, 50.2\%, and 71.2\% for UniIR, SigLIP, and CLIP, respectively. The varying degrees of reduction suggest that the impact of query diversification depends on the retriever, with CLIP retaining the highest attack success rate. Nevertheless, Poisoned-RAG accuracy remains consistently below Clean-RAG accuracy across all configurations, indicating that multi-query retrieval can mitigate but does not eliminate the influence of poisoned records.

\textbf{Category-shared attack under image-only retrieval.}
A more challenging setting is to reuse the same poisoned images across multiple queries rather than optimize them individually. \cref{apptab:category_image} shows that this category-shared setting achieves average ASRs of 28.6\%, 30.7\%, and 43.2\% for UniIR, SigLIP, and CLIP, respectively. Although the attack is less effective than its fixed-query counterpart, it still consistently reduces answer accuracy across all retrievers and VLMs. This demonstrates that poisoned images optimized for shared categories retain sufficient retrieval competitiveness across different queries within those categories, providing a degree of generalization without query-specific image optimization.

\begin{table*}[h]
\centering
\small
\caption{Fixed-query attack results under image-only retrieval (\%). N-Acc., C-Acc., and P-Acc. denote No-RAG, Clean-RAG, and Poisoned-RAG accuracy, respectively.}
\label{apptab:fixed_query_image}
\begin{tabular*}{\textwidth}{@{\extracolsep{\fill}}llc*{3}{ccc}@{}}
\toprule
\multirow{2}{*}{Scoring} &
\multirow{2}{*}{VLM} &
\multirow{2}{*}{\shortstack{N-Acc.}} &
\multicolumn{3}{c}{UniIR} &
\multicolumn{3}{c}{SigLIP} &
\multicolumn{3}{c}{CLIP} \\
\cmidrule(lr){4-6}\cmidrule(lr){7-9}\cmidrule(lr){10-12}
& & & C-Acc. & P-Acc. & ASR
& C-Acc. & P-Acc. & ASR
& C-Acc. & P-Acc. & ASR \\
\midrule
\multirow{4}{*}{Rule}
& LLaVA-1.6    & 8.7  & 30.4 & 4.4  & 95.9 & 34.2 & 3.3  & 98.6 & 22.5 & 3.9  & 98.5 \\
& Qwen2.5-VL   & 20.4 & 29.5 & 17.1 & 89.2 & 32.7 & 16.7 & 91.1 & 24.4 & 16.6 & 88.8 \\
& InternVL3.5  & 9.4  & 30.6 & 7.0  & 95.9 & 35.3 & 7.8  & 94.6 & 24.8 & 5.9  & 98.0 \\
& DeepSeek-VL2 & 15.8 & 26.0 & 9.3  & 95.7 & 31.2 & 9.5  & 96.4 & 20.2 & 8.8  & 97.1 \\
\midrule
\multirow{4}{*}{LLM}
& LLaVA-1.6    & 14.6 & 36.8 & 16.7 & 83.8 & 41.9 & 13.8 & 89.4 & 26.4 & 15.3 & 84.1 \\
& Qwen2.5-VL   & 28.5 & 38.9 & 30.6 & 74.2 & 44.3 & 30.6 & 77.7 & 35.0 & 30.1 & 72.6 \\
& InternVL3.5  & 18.8 & 37.0 & 18.1 & 85.8 & 43.0 & 16.9 & 88.1 & 29.2 & 16.1 & 89.9 \\
& DeepSeek-VL2 & 22.8 & 35.2 & 16.8 & 90.9 & 39.4 & 17.2 & 92.4 & 29.0 & 16.6 & 86.2 \\
\midrule
\multicolumn{2}{l}{\textbf{Average}}
& \textbf{17.4}
& \textbf{33.1} & \textbf{15.0} & \textbf{88.9}
& \textbf{37.8} & \textbf{14.5} & \textbf{91.0}
& \textbf{26.4} & \textbf{14.2} & \textbf{89.4} \\
\bottomrule
\end{tabular*}
\end{table*}
\begin{table*}[h]
\centering
\small
\caption{Fixed-query attack results under Multi-Query retrieval (\%). N-Acc., C-Acc., and P-Acc. denote No-RAG, Clean-RAG, and Poisoned-RAG accuracy, respectively.}
\label{apptab:fixed_query_multiquery}
\begin{tabular*}{\textwidth}{@{\extracolsep{\fill}}llc*{3}{ccc}@{}}
\toprule
\multirow{2}{*}{Scoring} &
\multirow{2}{*}{VLM} &
\multirow{2}{*}{\shortstack{N-Acc.}} &
\multicolumn{3}{c}{UniIR} &
\multicolumn{3}{c}{SigLIP} &
\multicolumn{3}{c}{CLIP} \\
\cmidrule(lr){4-6}\cmidrule(lr){7-9}\cmidrule(lr){10-12}
& & & C-Acc. & P-Acc. & ASR
& C-Acc. & P-Acc. & ASR
& C-Acc. & P-Acc. & ASR \\
\midrule
\multirow{4}{*}{Rule}
& LLaVA-1.6    & 8.7  & 36.1 & 18.7 & 57.5 & 30.0 & 20.7 & 43.5 & 11.6 & 6.6  & 77.4 \\
& Qwen2.5-VL   & 20.4 & 37.1 & 23.7 & 61.1 & 27.7 & 21.8 & 52.5 & 22.5 & 15.1 & 64.7 \\
& InternVL3.5  & 9.4  & 37.3 & 21.0 & 54.6 & 27.7 & 20.3 & 45.0 & 12.4 & 7.7  & 74.5 \\
& DeepSeek-VL2 & 15.8 & 31.8 & 19.6 & 62.1 & 25.0 & 18.4 & 55.9 & 17.7 & 9.7  & 67.4 \\
\midrule
\multirow{4}{*}{LLM}
& LLaVA-1.6    & 14.6 & 43.7 & 26.5 & 65.5 & 37.8 & 29.1 & 54.7 & 21.0 & 17.2 & 65.7 \\
& Qwen2.5-VL   & 28.5 & 46.6 & 37.2 & 54.0 & 39.6 & 33.4 & 52.1 & 34.2 & 27.7 & 62.8 \\
& InternVL3.5  & 18.8 & 45.4 & 31.2 & 52.7 & 36.8 & 31.7 & 41.7 & 25.2 & 15.9 & 78.4 \\
& DeepSeek-VL2 & 22.8 & 39.2 & 26.5 & 62.9 & 33.6 & 26.2 & 56.4 & 27.1 & 16.2 & 78.5 \\
\midrule
\multicolumn{2}{l}{\textbf{Average}}
& \textbf{17.4}
& \textbf{39.7} & \textbf{25.6} & \textbf{58.8}
& \textbf{32.3} & \textbf{25.2} & \textbf{50.2}
& \textbf{21.5} & \textbf{14.5} & \textbf{71.2} \\
\bottomrule
\end{tabular*}
\end{table*}
\begin{table*}[h]
\centering
\small
\caption{Category-shared attack results under image-only retrieval (\%). N-Acc., C-Acc., and P-Acc. denote No-RAG, Clean-RAG, and Poisoned-RAG accuracy, respectively.}
\label{apptab:category_image}
\begin{tabular*}{\textwidth}{@{\extracolsep{\fill}}llc*{3}{ccc}@{}}
\toprule
\multirow{2}{*}{Scoring} & \multirow{2}{*}{VLM} & \multirow{2}{*}{N-Acc.} &
\multicolumn{3}{c}{UniIR} & \multicolumn{3}{c}{SigLIP} & \multicolumn{3}{c}{CLIP} \\
\cmidrule(lr){4-6}\cmidrule(lr){7-9}\cmidrule(lr){10-12}
& & & C-Acc. & P-Acc. & ASR & C-Acc. & P-Acc. & ASR & C-Acc. & P-Acc. & ASR \\
\midrule
\multirow{4}{*}{Rule}
& LLaVA-1.6   & 11.4 & 33.6 & 20.8 & 31.9 & 39.1 & 25.0 & 34.8 & 19.2 & 11.2 & 45.1 \\
& Qwen2.5-VL  & 19.7 & 31.1 & 25.5 & 22.4 & 33.9 & 30.1 & 30.0 & 23.9 & 16.4 & 47.6 \\
& InternVL3.5 & 13.8 & 29.1 & 25.0 & 27.1 & 34.6 & 27.3 & 29.1 & 18.0 & 14.9 & 50.7 \\
& DeepSeek-VL2 & 18.9 & 26.8 & 21.9 & 28.6 & 31.5 & 26.8 & 29.9 & 24.2 & 14.6 & 47.8 \\
\midrule
\multirow{4}{*}{LLM}
& LLaVA-1.6   & 15.3 & 42.4 & 30.1 & 29.0 & 46.6 & 32.2 & 30.2 & 24.2 & 18.6 & 29.6 \\
& Qwen2.5-VL  & 25.7 & 41.8 & 35.0 & 30.0 & 44.9 & 41.8 & 28.8 & 33.2 & 26.8 & 43.1 \\
& InternVL3.5 & 17.0 & 39.1 & 31.1 & 28.1 & 45.5 & 35.0 & 30.3 & 24.3 & 20.8 & 29.1 \\
& DeepSeek-VL2 & 22.4 & 32.2 & 28.4 & 31.6 & 37.7 & 33.5 & 32.3 & 28.1 & 20.4 & 52.5 \\
\midrule
\multicolumn{2}{l}{\textbf{Average}}
& \textbf{18.0} & \textbf{34.5} & \textbf{27.2} & \textbf{28.6}
& \textbf{39.2} & \textbf{31.5} & \textbf{30.7}
& \textbf{24.4} & \textbf{18.0} & \textbf{43.2} \\
\bottomrule
\end{tabular*}
\end{table*}

\subsection{Human Validation of Rule/LLM Judge}
\label{appsec:human_validation}

\begin{table}[t]
\centering
\caption{Agreement between rule-based matching and human annotations.}
\label{apptab:human_validation_rule}
\begin{tabular}{lccc}
\toprule
Human Annotation & Correct & Incorrect & Total \\
\midrule
Correct   & 68  & 19  & 87  \\
Incorrect & 0   & 213 & 213 \\
\midrule
Total     & 68  & 232 & 300 \\
\bottomrule
\end{tabular}
\end{table}
\begin{table}[t]
\centering
\caption{Agreement between LLM Judge and human annotations.}
\label{apptab:human_validation_phi}
\begin{tabular}{lccc}
\toprule
Human Annotation & Correct & Incorrect & Total \\
\midrule
Correct   & 84  & 3   & 87  \\
Incorrect & 21  & 192 & 213 \\
\midrule
Total     & 105 & 195 & 300 \\
\bottomrule
\end{tabular}
\end{table}

In our experiments, we use rule-based matching and LLM judging to evaluate answer correctness. To further validate the reliability of these two evaluation methods, we conduct a human evaluation on 300 randomly sampled responses, including 100 responses each from No-RAG, Clean-RAG, and Poisoned-RAG. Human annotators independently assess each response based on the question, ground-truth answer, and model response, without knowing the retrieval setting or the automatic evaluation results. Responses that provide the ground-truth answer or a semantically equivalent answer are labeled as correct, while irrelevant, contradictory, or explicitly uncertain responses are labeled as incorrect.

As shown in \cref{apptab:human_validation_rule} and \cref{apptab:human_validation_phi}, rule-based matching achieves 93.7\% agreement with human annotations, with a Cohen's Kappa of 0.836. LLM judging achieves 92.0\% agreement, with a Cohen's Kappa of 0.817. These results demonstrate that both evaluation methods achieve strong agreement with human judgments, supporting their use in our large-scale evaluation.

\subsection{Random Seed Evaluation}

\begin{table*}[t]
\centering
\small
\setlength{\tabcolsep}{4pt}
\caption{Attack results under equal-weight image-text fusion with an alternative random seed (\%). Each paired value is reported as Rule/LLM. N-Acc., C-Acc., and P-Acc. denote No-RAG, Clean-RAG, and Poisoned-RAG accuracy, respectively.}
\label{apptab:random_seed}
\begin{tabular*}{\textwidth}{@{\extracolsep{\fill}}lc*{3}{ccc}@{}}
\toprule
\multirow{2}{*}{VLM} &
\multirow{2}{*}{N-Acc.} &
\multicolumn{3}{c}{UniIR} &
\multicolumn{3}{c}{SigLIP} &
\multicolumn{3}{c}{CLIP} \\
\cmidrule(lr){3-5}\cmidrule(lr){6-8}\cmidrule(lr){9-11}
& & C-Acc. & P-Acc. & ASR
& C-Acc. & P-Acc. & ASR
& C-Acc. & P-Acc. & ASR \\
\midrule

\multicolumn{11}{l}{\textbf{Fixed-query}} \\
\addlinespace[2pt]

LLaVA-1.6 & 9.0/17.4
& 33.1/38.8 & 7.2/14.9 & 92.7/89.6
& 31.3/38.1 & 13.9/25.9 & 71.6/68.9
& 18.4/25.9 & 7.5/17.7 & 82.8/70.2 \\

Qwen2.5-VL & 19.1/28.7
& 34.0/42.3 & 16.3/29.2 & 84.3/81.9
& 25.8/38.3 & 20.9/31.1 & 70.2/62.3
& 24.8/34.4 & 17.0/31.2 & 63.8/65.3 \\

InternVL3.5 & 9.3/19.3
& 30.7/41.8 & 8.1/19.3 & 90.8/82.0
& 29.1/36.8 & 16.4/29.6 & 71.0/64.5
& 16.5/22.4 & 4.9/13.8 & 82.3/81.0 \\

DeepSeek-VL2 & 16.0/22.4
& 29.0/39.0 & 9.3/18.3 & 99.6/87.9
& 22.9/36.3 & 16.3/27.1 & 70.1/69.0
& 18.1/26.4 & 10.0/17.8 & 80.3/80.9 \\

\midrule
\textbf{Average} & \textbf{13.4/22.0}
& \textbf{31.7/40.5} & \textbf{10.2/20.4} & \textbf{91.9/85.4}
& \textbf{27.3/37.4} & \textbf{16.9/28.4} & \textbf{70.7/66.2}
& \textbf{19.5/27.3} & \textbf{9.8/20.1} & \textbf{77.3/74.4} \\

\midrule
\multicolumn{11}{l}{\textbf{Category-shared}} \\
\addlinespace[2pt]

LLaVA-1.6 & 11.4/13.9
& 34.7/39.6 & 25.1/32.8 & 66.8/65.5
& 36.8/46.2 & 32.8/40.4 & 35.5/39.1
& 29.3/32.7 & 25.7/26.1 & 44.0/44.1 \\

Qwen2.5-VL & 21.8/24.7
& 37.0/44.1 & 34.5/38.8 & 63.6/54.0
& 37.0/44.7 & 32.2/39.3 & 36.7/43.0
& 34.9/40.3 & 28.8/36.8 & 51.6/37.9 \\

InternVL3.5 & 13.6/18.0
& 37.0/39.9 & 28.9/29.5 & 74.5/73.9
& 40.1/44.2 & 36.9/43.6 & 34.7/29.9
& 28.6/34.3 & 23.0/26.9 & 43.7/47.8 \\

DeepSeek-VL2 & 18.4/22.6
& 35.7/42.3 & 25.8/31.2 & 74.9/69.3
& 36.9/44.7 & 34.2/39.5 & 36.2/46.3
& 31.8/36.6 & 26.5/30.1 & 53.3/48.1 \\

\midrule
\textbf{Average} & \textbf{16.3/19.8}
& \textbf{36.1/41.5} & \textbf{28.6/33.1} & \textbf{70.0/65.7}
& \textbf{37.7/45.0} & \textbf{34.0/40.7} & \textbf{35.8/39.6}
& \textbf{31.1/36.0} & \textbf{26.0/30.0} & \textbf{48.2/44.5} \\

\bottomrule
\end{tabular*}
\end{table*}

We evaluate FAIRP on a query subset from InfoSeek Task 8 in the main experiments. To further investigate the sensitivity of the results to random sampling, we change the random seed for query sampling and rerun the experiments while keeping all other experimental configurations unchanged. As shown in \cref{apptab:random_seed}, both settings maintain substantial ASRs across the three retrievers and two evaluation methods, with overall trends consistent with those of the main experiments. The changes in mean ASR relative to the main experiments are only 0.05 and 0.97 percentage points for Fixed-query and Category-shared, respectively. This demonstrates that the effectiveness of FAIRP and its main conclusions do not depend on a particular query sample. Poisoned images are directly initialized from their corresponding source images rather than random noise.
\section{Attack Robustness}
\label{appsec:robust}

\subsection{Question Reformulation}

\begin{figure*}[t]
    \centering
    \includegraphics[width=\textwidth]{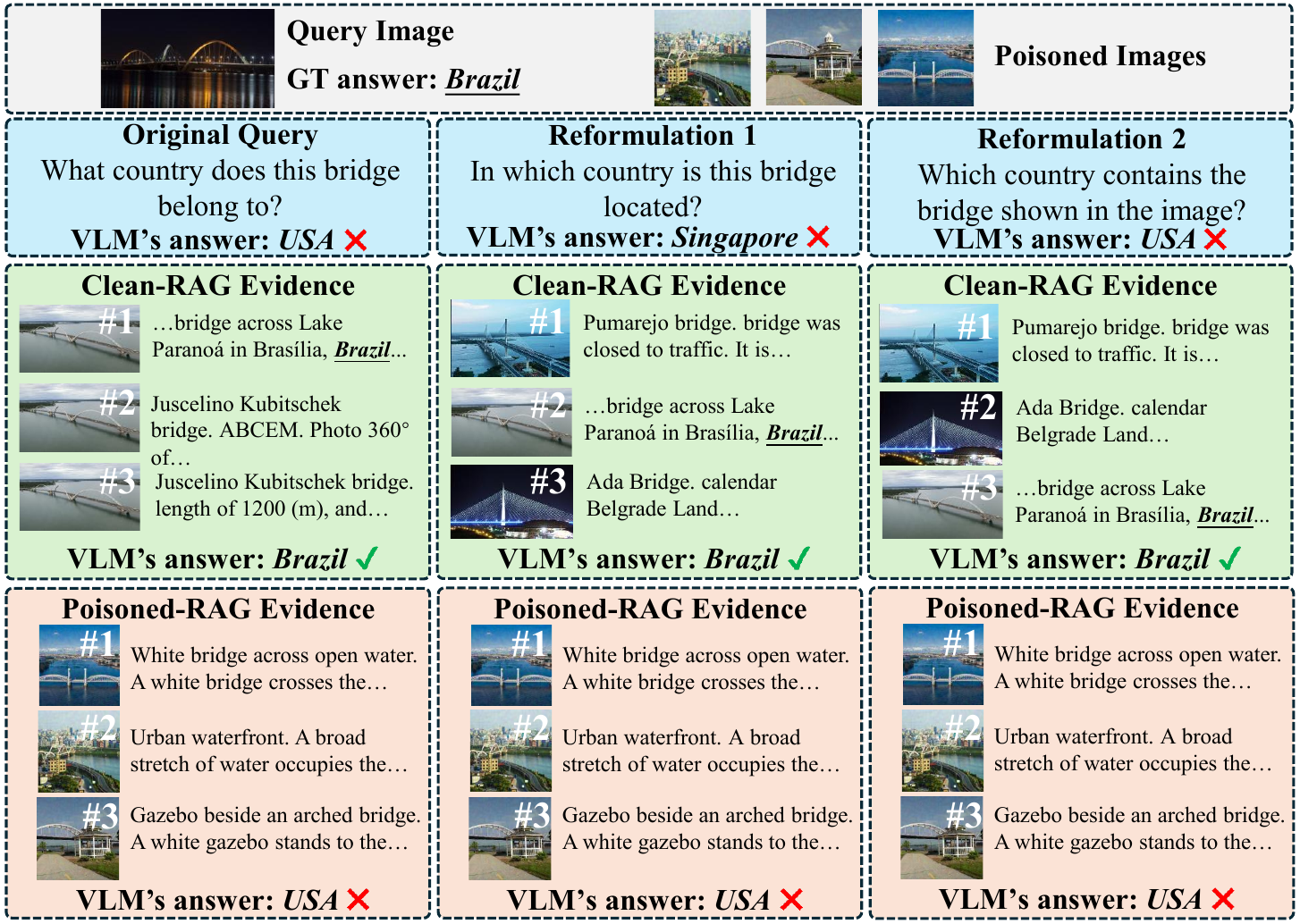}
    \caption{Qualitative example of attack retention under question reformulation. Clean-RAG correctly answers the original query and both reformulations with evidence identifying Brazil. In contrast, reusing the same Poisoned Images consistently leads to incorrect USA answers, without re-optimization.}
    \label{appfig:question_reformulation}
\end{figure*}

In \cref{sec:question_reformulation} of the main paper, we quantitatively evaluate the robustness of FAIRP against question reformulation using Attack Retention Rate (ARR), achieving an average ARR of 84.4\%. Here, \cref{appfig:question_reformulation} further presents a representative example to illustrate this process.

For the original query and its two semantically equivalent reformulations, Clean-RAG consistently retrieves evidence containing the correct answer and enables the VLM to answer correctly. In contrast, when reusing the same poisoned images, the poisoned records enter the Top-3 results under all three question formulations and lead to incorrect VLM answers. This demonstrates that FAIRP maintains its attack effectiveness despite changes in query wording, without re-optimizing the images for reformulated queries.

\subsection{Evaluation with more $K$ and $P$}

\begin{figure*}[t]
    \centering
    \begin{minipage}{\textwidth}
        \centering
        \includegraphics[width=\textwidth]{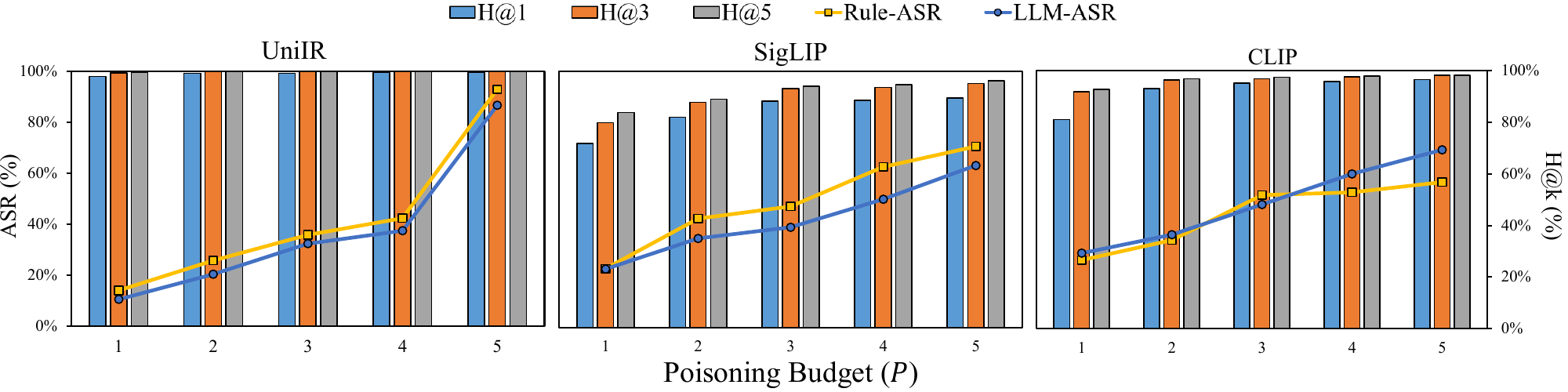}
        \centerline{\small (a) Fixed-query}
    \end{minipage}

    \vspace{3pt}

    \begin{minipage}{\textwidth}
        \centering
        \includegraphics[width=\textwidth]{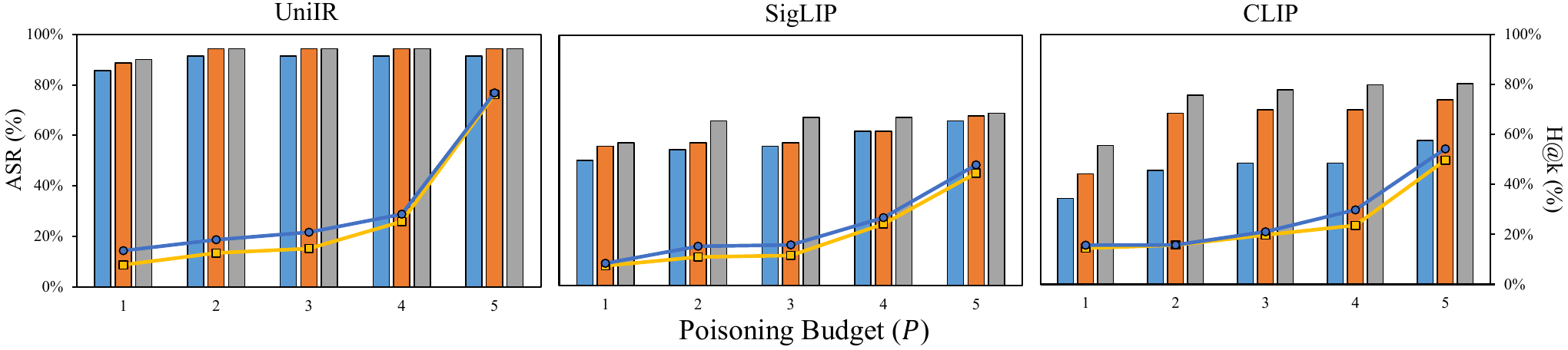}
        \centerline{\small (b) Category-shared}
    \end{minipage}

    \caption{Effect of poisoning budget $P$ at $K=5$ under (a) Fixed-query and (b) Category-shared settings. Bars represent retrieval hit rates (H@1, H@3, and H@5), while lines represent Rule-ASR and LLM-ASR.}
    \label{appfig:kp_k5}
\end{figure*}
\begin{table}[t]
\centering
\small
\setlength{\tabcolsep}{5pt}
\caption{Results at $K=1$ under Fixed-query and Category-shared settings (\%).}
\label{apptab:kp_k1}
\begin{tabular}{llccc}
\toprule
Setting & Retriever & H@1 & Rule-ASR & LLM-ASR \\
\midrule
\multirow{3}{*}{Fixed-query}
& UniIR  & 98.1 & 98.0 & 80.5 \\
& SigLIP & 71.9 & 48.3 & 41.6 \\
& CLIP   & 81.1 & 54.3 & 62.3 \\
\midrule
\multirow{3}{*}{Category-shared}
& UniIR  & 85.7 & 78.2 & 70.8 \\
& SigLIP & 48.1 & 31.3 & 30.3 \\
& CLIP   & 35.2 & 30.5 & 28.6 \\
\bottomrule
\end{tabular}
\end{table}

In the main experiments, we use $K=3$ by default and evaluate the impact of different Poisoning Budgets $P$ on attack effectiveness (\cref{sec:ablation}). To further investigate the attack performance under different retrieval depths, we additionally evaluate $K=1$ and $K=5$ under both Fixed-query and Category-shared settings, while keeping all other experimental configurations unchanged.

When $K=1$, a single poisoned image is already sufficient to affect Top-1 retrieval and downstream VLM answers. We report $H@1$ and ASR for all three retrievers under both settings in \cref{apptab:kp_k1}. Overall, Fixed-query achieves better attack performance than Category-shared. In terms of ASR, UniIR achieves the strongest results under both settings, suggesting that the effectiveness of a single poisoned image is sensitive to differences in retriever representations.

When $K=5$, increasing the Poisoning Budget further improves attack effectiveness. As shown in \cref{appfig:kp_k5}, under both Fixed-query and Category-shared settings, the retrieval hit rates of all three retrievers generally increase or saturate as $P$ increases, while ASR also exhibits an overall increasing trend. In particular, ASR increases substantially at $P=5$ under both settings. This indicates that, with a larger retrieval depth, injecting more poisoned images can increase their influence on retrieval results and further improve their ability to hijack downstream VLM answers. Unlike existing methods that manipulate and inject malicious text, our method does not require control over the associated text. Instead, it causes seemingly harmless but irrelevant evidence to enter the Top-K results, preventing the retriever from providing useful evidence for VLM answer generation. Consequently, a larger $P$ is naturally required as $K$ increases, and the attack is more effective when poisoned images occupy more positions in the Top-K evidence.

\subsection{Image Preprocessing}

\begin{figure*}[t]
    \centering
    % First row: Random perturbation baselines
    \begin{subfigure}[t]{0.24\textwidth}
        \centering
        \includegraphics[width=\linewidth]{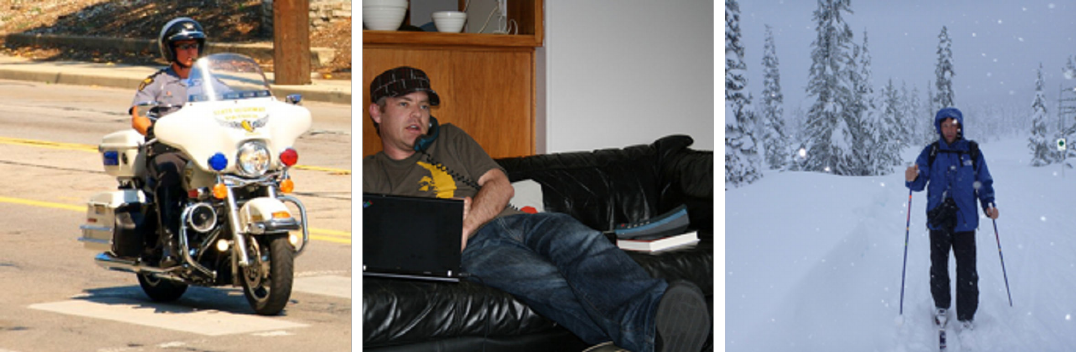}
        \caption{Source Images}
    \end{subfigure}
    \hfill
    \begin{subfigure}[t]{0.24\textwidth}
        \centering
        \includegraphics[width=\linewidth]{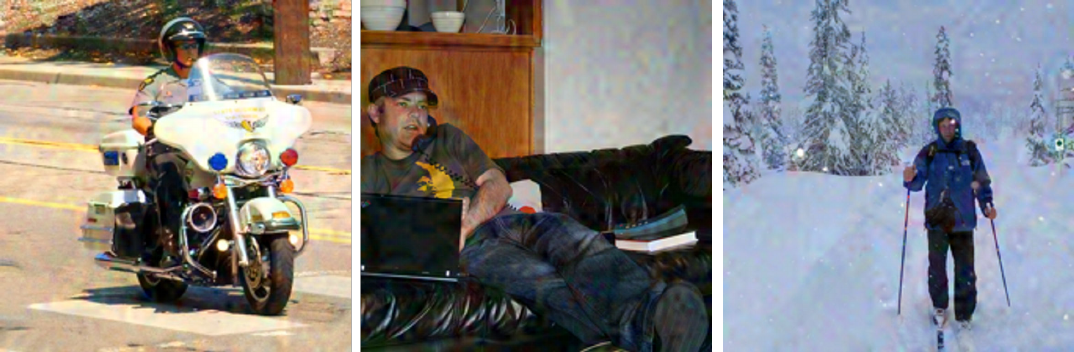}
        \caption{FAIRP Poisoned Images (Ours)}
    \end{subfigure}
    \hfill
    \begin{subfigure}[t]{0.24\textwidth}
        \centering
        \includegraphics[width=\linewidth]{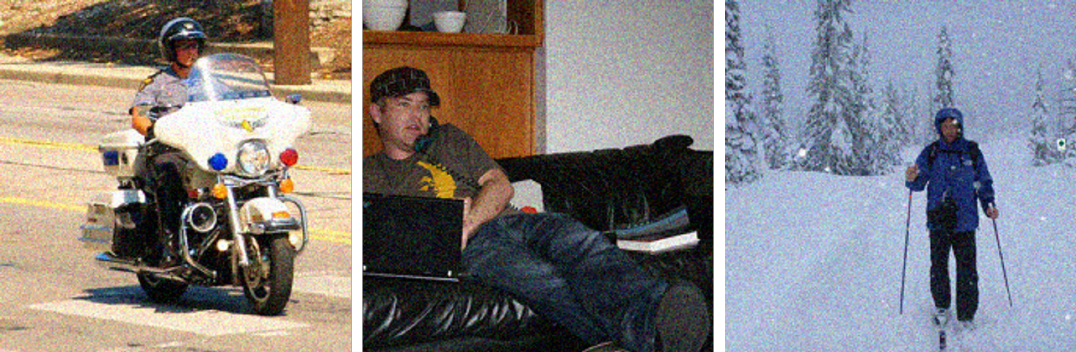}
        \caption{Strong Random Perturbations}
    \end{subfigure}
    \hfill
    \begin{subfigure}[t]{0.24\textwidth}
        \centering
        \includegraphics[width=\linewidth]{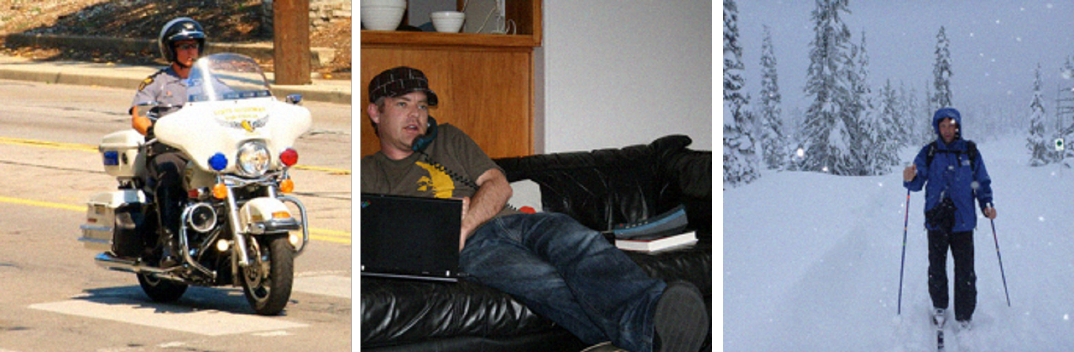}
        \caption{Weak Random Perturbations}
    \end{subfigure}

    \vspace{6pt}

    % Second row: Image preprocessing
    \begin{subfigure}[t]{0.24\textwidth}
        \centering
        \includegraphics[width=\linewidth]{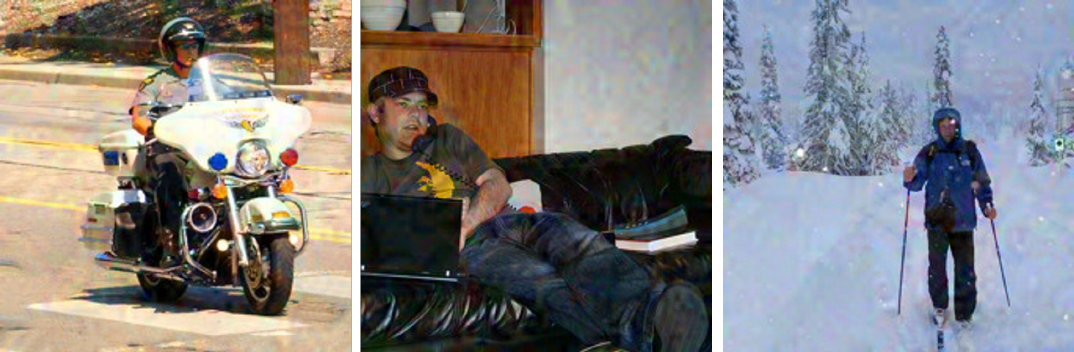}
        \caption{JPEG Compression (Q=85)}
    \end{subfigure}
    \hfill
    \begin{subfigure}[t]{0.24\textwidth}
        \centering
        \includegraphics[width=\linewidth]{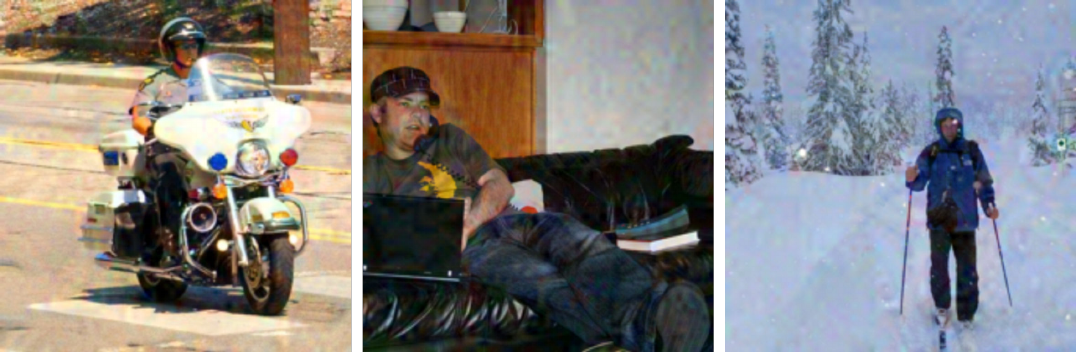}
        \caption{Gaussian Blur ($\sigma=0.75$)}
    \end{subfigure}
    \hfill
    \begin{subfigure}[t]{0.24\textwidth}
        \centering
        \includegraphics[width=\linewidth]{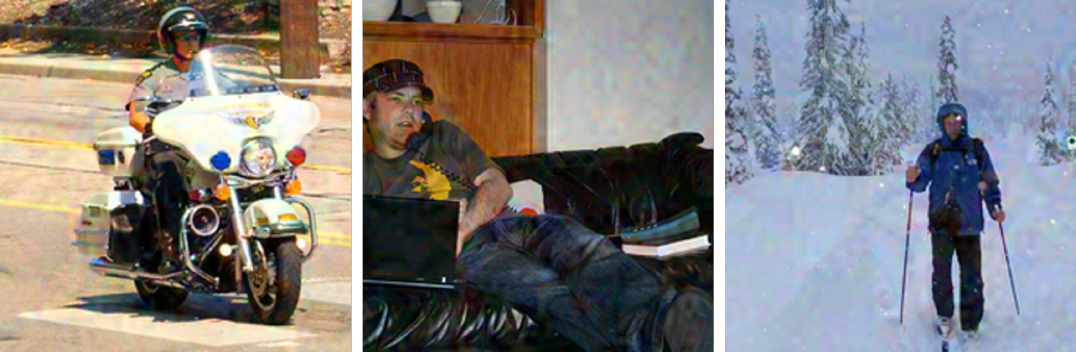}
        \caption{3\% Cropping}
    \end{subfigure}
    \hfill
    \begin{subfigure}[t]{0.24\textwidth}
        \centering
        \includegraphics[width=\linewidth]{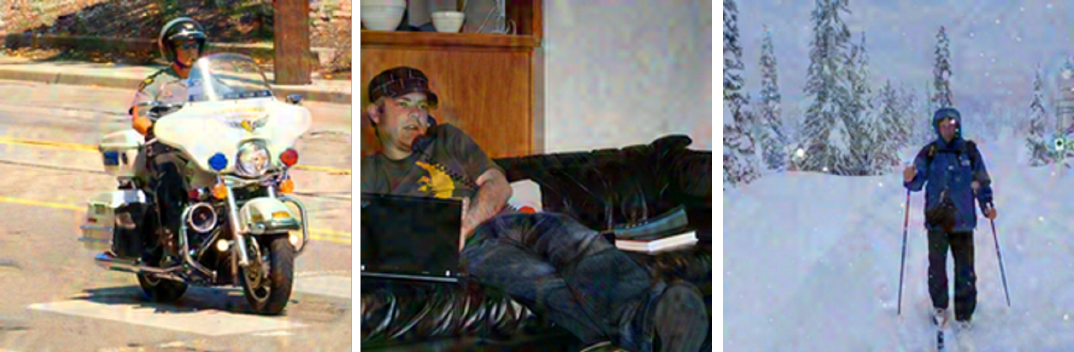}
        \caption{90\% Resizing}
    \end{subfigure}

    \caption{Visual comparisons of source images, random perturbations, FAIRP-optimized poisoned images, and common image preprocessing operations.}
    \label{appfig:visual_comparisons}
\end{figure*}

Following the experimental setting in the main paper, the attacker submits poisoned images to the KB's normal ingestion system, which then performs subsequent processing and retrieval. To evaluate the robustness of FAIRP against common image preprocessing operations, we select 100 originally successful queries based on the LLM Judge results in the main experiments. According to the ASR definition in the main paper, a query is considered a successful attack only when all four conditions are satisfied: the No-RAG answer is incorrect, the Clean-RAG answer is correct, at least one poisoned record enters the Top-$K$ retrieved evidence, and the Poisoned-RAG answer is incorrect. We apply four preprocessing operations to the corresponding poisoned images: JPEG compression (Quality=85), Gaussian blur ($\sigma=0.75$), cropping 3\% of the image boundaries followed by resizing to the original dimensions, and downscaling to 90\% of the original resolution followed by restoration to the original dimensions. The visual results are shown in \cref{appfig:visual_comparisons} (e)-(h). The preprocessed images are not re-optimized and are evaluated using the same configurations as the main experiments.

The results show that, under the four preprocessing operations, 98, 97, 98, and 79 queries, respectively, still have their preprocessed poisoned images retrieved in the Top-3, while 93, 95, 96, and 74 queries retain attack success. These results demonstrate that FAIRP exhibits strong robustness to JPEG compression, Gaussian blur, and mild cropping, whereas image resizing has a somewhat more noticeable impact. This may be because downscaling and subsequently restoring images to their original dimensions introduce substantial interpolation changes, thereby affecting the optimized visual feature representations. Nevertheless, most originally successful queries retain their attack effectiveness after preprocessing, demonstrating the considerable robustness of FAIRP to common image preprocessing operations.
\section{Additional Comparison with Baselines}
\label{appsec:baseline}

\subsection{Simple Image Injection Baselines}

\begin{figure*}[t]
    \centering
    \includegraphics[width=0.95\textwidth,height=0.85\textheight,keepaspectratio]{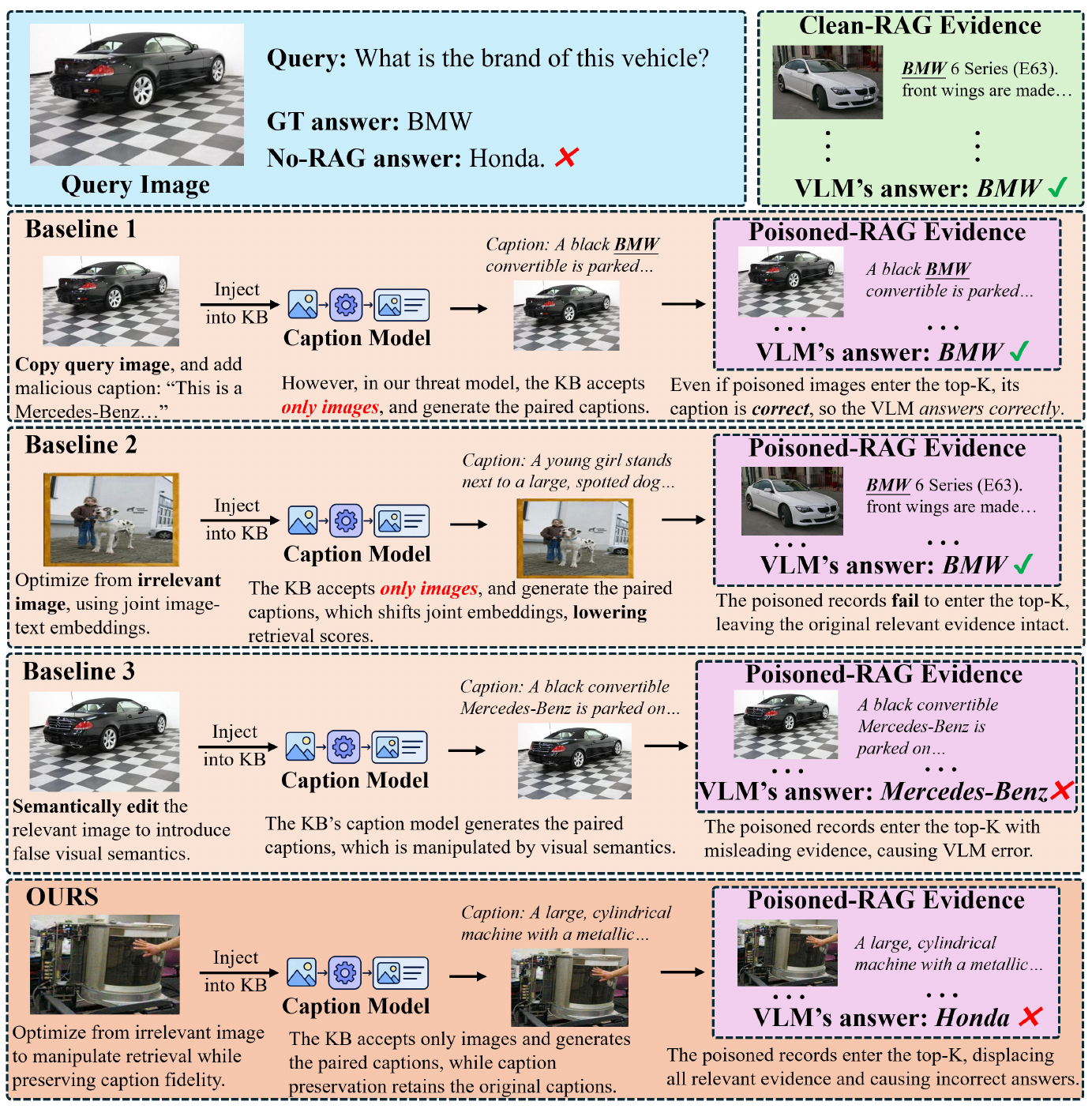}
    \caption{Comparison of existing multimodal RAG poisoning baselines and FAIRP under the image-only KB ingestion constraint.}
    \label{appfig:baseline_comparison}
\end{figure*}

To examine whether the attack effectiveness of FAIRP merely results from injecting images into the KB or applying generic pixel perturbations, we select three representative samples and visualize their source images, weak random perturbations, strong random perturbations, and FAIRP-optimized poisoned images in \cref{appfig:visual_comparisons} (a)-(d).

We further evaluate the source images and the weak and strong random perturbation baselines on the same 1,000 queries used in the main experiments, following exactly the same Fixed-query configurations. All three variants achieve an H@3 and ASR of 0. This demonstrates that image optimization is essential for FAIRP to achieve effective attack, as neither unmodified source images nor random perturbations can make poisoned records competitive in retrieval.

\subsection{Visual Comparison with Existing Attacks}
\textbf{Experimental Setup.}
We consider representative image construction, image optimization, and joint image-text retrieval strategies from existing multimodal RAG poisoning methods, including PoisonedEye~\cite{zhang2025poisonedeye}, MM-PoisonRAG~\cite{ha-etal-2026-mm}, and M³Att~\cite{yang2026knowledge}. These methods mainly adopt two image construction strategies: one directly copies or selects images relevant to the query and combines them with attacker-crafted malicious text; the other selects irrelevant source images, while also controlling the paired captions, and optimizes image perturbations through joint image-text embeddings to improve the retrieval competitiveness of poisoned records under image-text fusion.

We adapt these strategies to our threat model: the attacker can construct or optimize poisoned images following the corresponding methods but cannot control the paired text actually stored in the KB. During injection, the attacker can only submit images, and the KB automatically generates paired captions using its normal captioning model before adding the resulting image-text records to the original KB.

We use the same Fixed-query setting and Equal-weight Image-text Fusion configuration as in the main experiments, ensuring that the evaluation protocol is fully consistent with the main experiments and that the comparison is fair.

\textbf{Experimental Results and Analysis.}
As discussed in \cref{sec:baseline} of the main paper, existing multimodal RAG poisoning methods have limitations under the image-only KB ingestion constraint. Here, we further demonstrate their failure mechanisms through specific experiments and case studies.

\begin{figure*}[t]
    \centering
    \includegraphics[width=\textwidth]{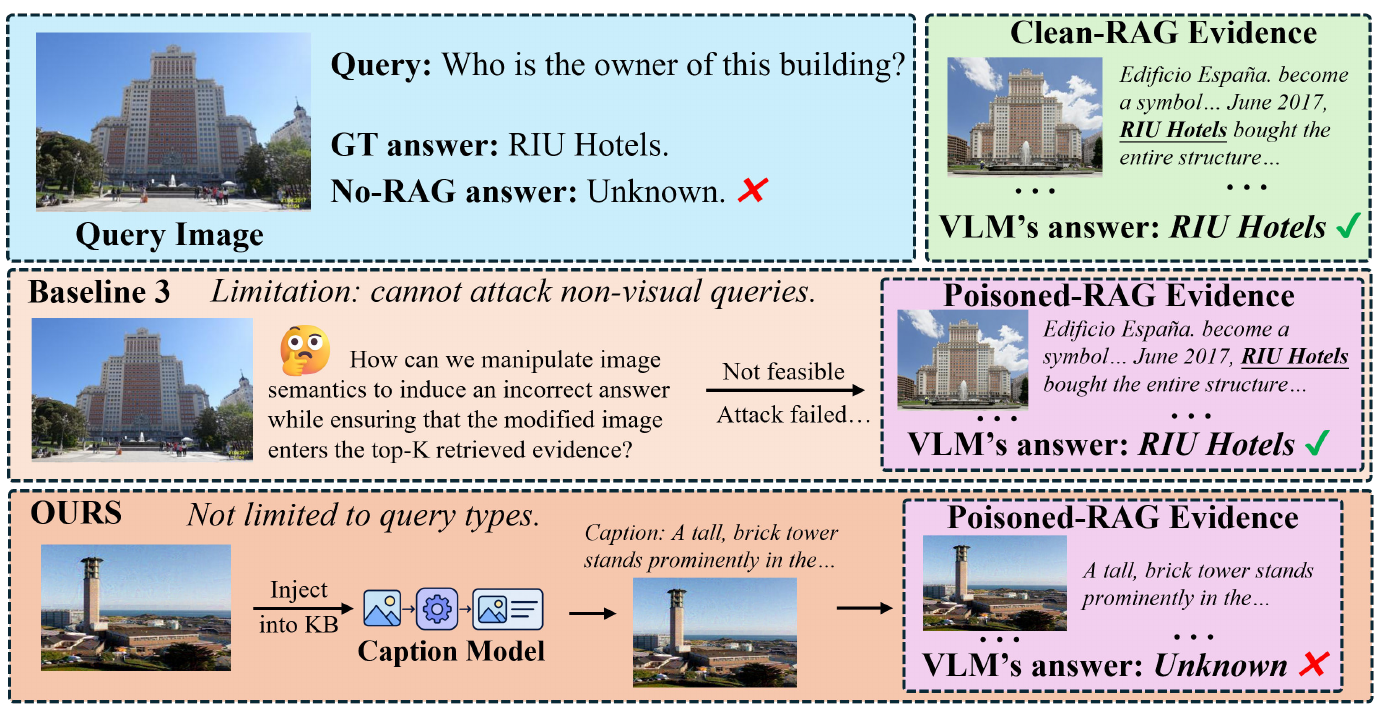}
    \caption{Comparison of Vis-Poison and FAIRP on non-visual queries. Vis-Poison cannot effectively attack non-visual queries, whereas FAIRP is not restricted to specific query types.}
    \label{appfig:non_visual_comparison}
\end{figure*}

Depending on the semantic relevance between poisoned images and queries, as well as their construction strategies, we observe two distinct attack failure cases, as illustrated in \cref{appfig:baseline_comparison}. For clarity and ease of comparison, each method is illustrated with only one representative poisoned image, while the remaining poisoned images are omitted from the figure.

In the first case (Baseline 1), the attacker directly copies the query image or selects an image relevant to the query, attempting to induce incorrect answers through malicious text. However, since the KB only accepts images and automatically generates captions, the resulting captions still correctly describe the image content. Even if the poisoned records successfully enter the Top-3, they cannot provide the attacker-intended malicious semantics. Consequently, the VLM can still answer correctly based on the relevant evidence.

In the second case (Baseline 2), the attacker selects irrelevant source images and optimizes image perturbations using attacker-crafted malicious text through joint image-text embeddings. Specifically, the perturbation optimization is based on embeddings fused from images and malicious text. However, during actual ingestion, the automatically generated captions from the KB replace the malicious text used during optimization. This causes a substantial decrease in the final joint embeddings of the images and their newly generated captions, reducing the retrieval competitiveness of poisoned records and preventing them from entering the Top-3. Consequently, the original relevant evidence remains in the retrieved results.

Across the 1,000 queries, these two baselines achieve an average ASR of only 2.3\%. This demonstrates that, without control over the paired text, existing poisoning strategies relying on image relevance or malicious joint image-text optimization struggle to achieve effective end-to-end answer hijacking.

\textbf{Comparison with Vis-Poison.}
\cref{sec:baseline} of the main paper also discusses poisoning methods based on visual semantic manipulation. Here, we further analyze their attack applicability and computational costs and compare them with FAIRP.

We select Vis-Poison~\cite{liang2026vis} as a representative method. It directly modifies the visual semantics of images relevant to the query to introduce misleading visual content, after which the KB's captioning model generates the corresponding descriptions. When the modified images successfully enter the Top-K, the resulting misleading image-text evidence can indeed induce incorrect VLM answers, as illustrated by Baseline 3 in \cref{appfig:baseline_comparison}.

However, compared with FAIRP, this type of method has two limitations.

First, as illustrated in \cref{appfig:non_visual_comparison}, its attack relies on targeted modifications to image semantics and is therefore primarily applicable to queries with explicit visual semantics. It cannot directly attack non-visual questions, such as questions about building ownership. Notably, 801 of the 1,000 queries used in our experiments are non-visual questions, which substantially limits the applicability of such methods.

Second, attackers need to design and optimize misleading image semantics for each target query while ensuring that the modified images enter the Top-K, requiring additional time and computational resources. We measure the time and computational costs of Vis-Poison and FAIRP on 100 visual queries. The total cost of Vis-Poison is approximately 3.53 times that of FAIRP. It is worth noting that this experiment uses only simplified versions of its planner and verifier, without fully deploying its large-scale planner and verifier, which would incur even higher costs.

In contrast, FAIRP does not need to modify the original image semantics. Instead, it optimizes the visual representations of irrelevant source images to achieve high retrieval competitiveness while preserving the fidelity of automatically generated captions. Therefore, even for non-visual queries that cannot be directly attacked through visual semantic manipulation, FAIRP can still interfere with correct answers from downstream VLMs by introducing irrelevant poisoned records into the Top-K and displacing the original relevant evidence. This demonstrates that FAIRP is not restricted to specific query types and has broader attack applicability.
\section{Additional Results of Caption Preservation}
\label{appsec:caption}

\begin{figure*}[t]
    \centering
    \begin{subfigure}[t]{\textwidth}
        \centering
        \includegraphics[width=\textwidth]{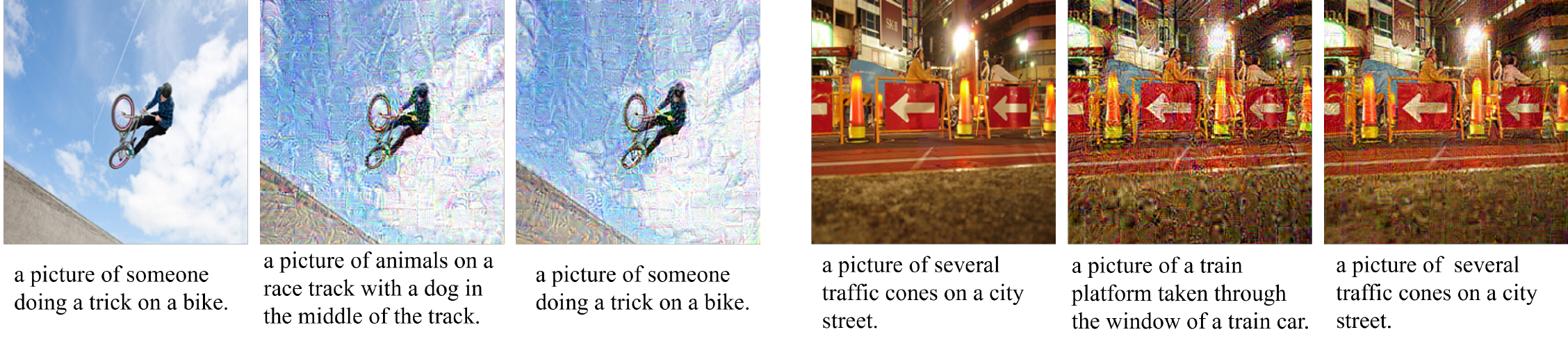}
        \caption{BLIP}
    \end{subfigure}

    \vspace{5pt}

    \begin{subfigure}[t]{\textwidth}
        \centering
        \includegraphics[width=\textwidth]{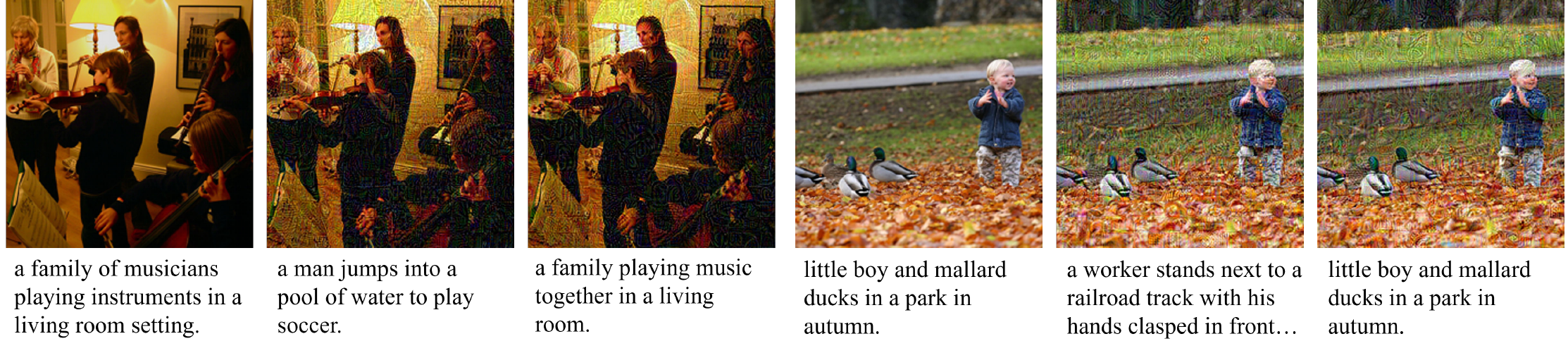}
        \caption{CoCa}
    \end{subfigure}

    \caption{Qualitative comparison of caption preservation loss. The top and bottom subfigures show BLIP and CoCa results, respectively. Each example presents the source image and the optimized images without and with caption preservation loss, together with their generated captions. The style loss is disabled in this experiment.}
    \label{appfig:caption_preservation}
\end{figure*}
\begin{figure*}[t]
    \centering

    \begin{subfigure}[t]{0.6\textwidth}
        \centering
        \includegraphics[width=\linewidth]{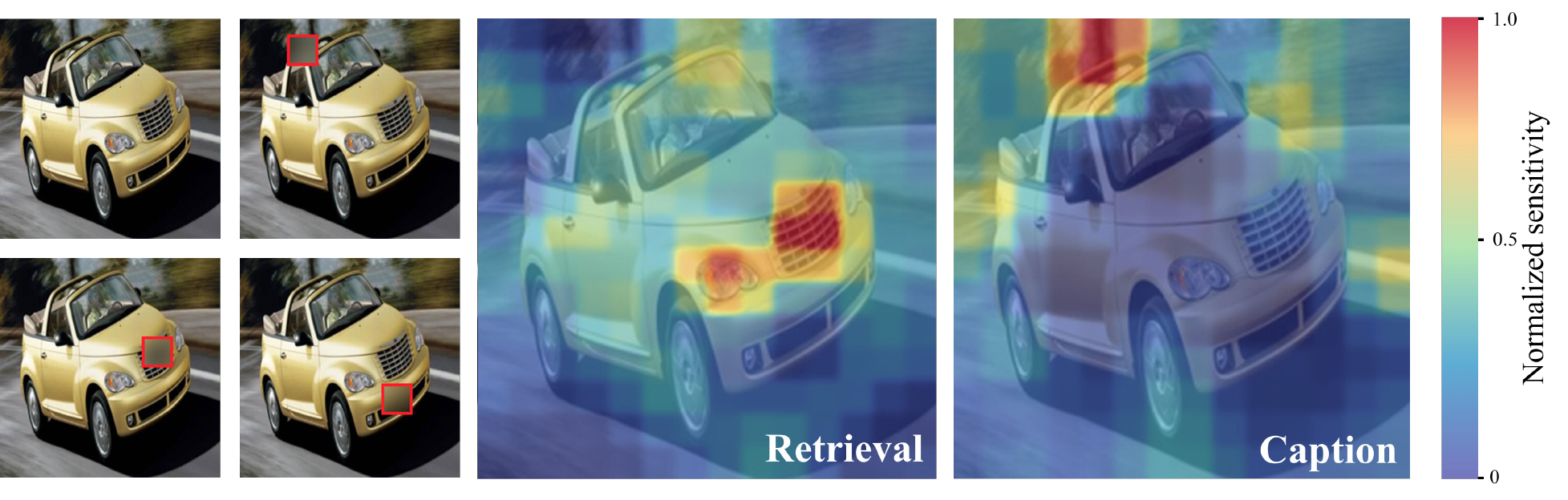}
        \caption{Query: What equipment or engine is used by this vehicle to provide power?}
        \label{appfig:visual_sensitivity_a}
    \end{subfigure}

    \vspace{4pt}

    \begin{subfigure}[t]{0.6\textwidth}
        \centering
        \includegraphics[width=\linewidth]{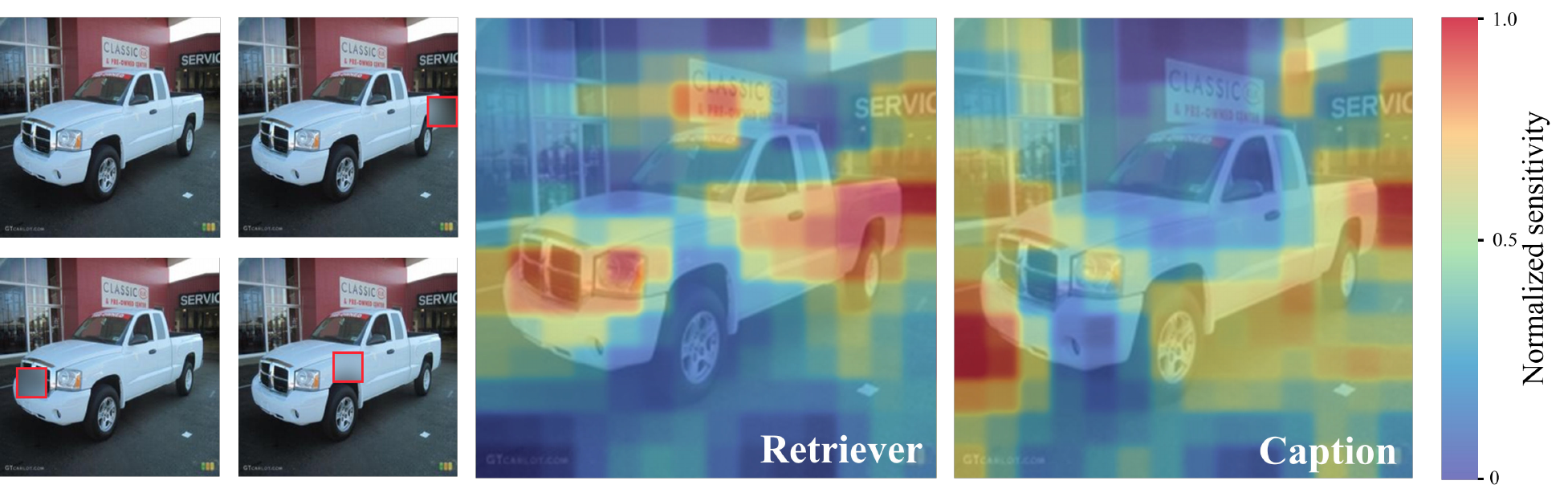}
        \caption{Query: What is the brand of this vehicle?}
        \label{appfig:visual_sensitivity_b}
    \end{subfigure}

    \vspace{4pt}

    \begin{subfigure}[t]{0.6\textwidth}
        \centering
        \includegraphics[width=\linewidth]{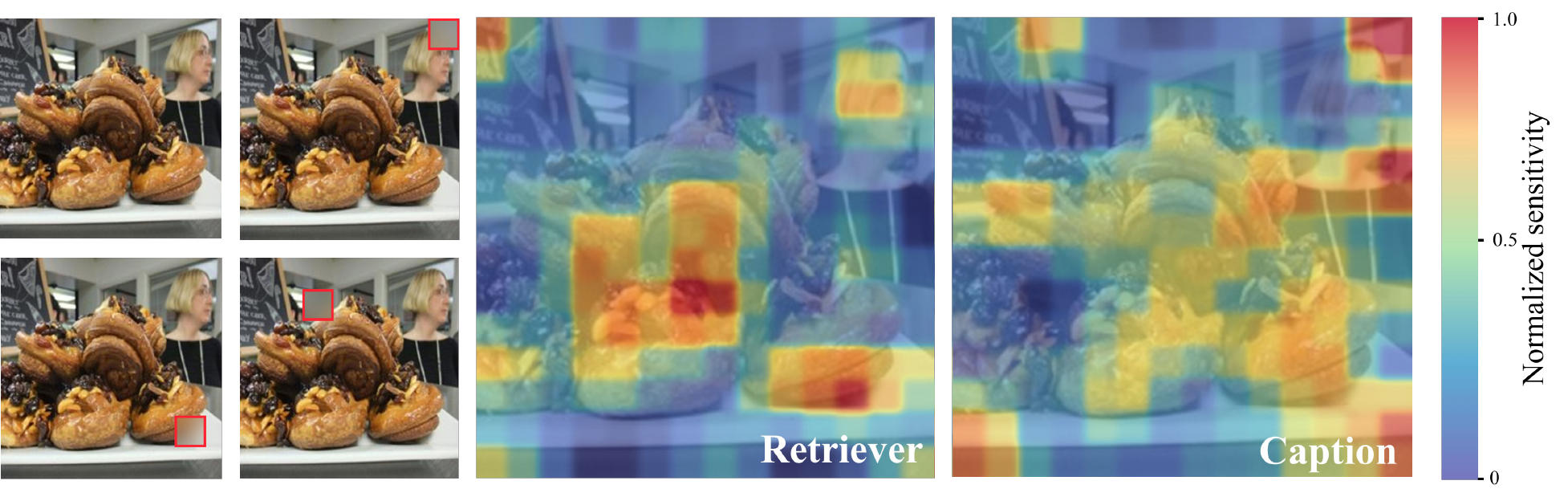}
        \caption{Query: Which country does this food come from?}
        \label{appfig:visual_sensitivity_c}
    \end{subfigure}

    \caption{Examples of visual sensitivity analysis. Each subfigure includes the original query image, three locally occluded images, and the corresponding retrieval and caption sensitivity maps. The high-sensitivity regions of the two tasks partially overlap but are not fully aligned, indicating that retrieval and caption generation rely on both shared and task-specific local visual information.}
    
    \label{appfig:map}
\end{figure*}

\textbf{Caption Retention Rate.}
In the main paper, we use the Caption Retention Rate (CRR) to measure the preservation of original caption outputs after image optimization. In this section, we provide its formal definition. Specifically, CRR is defined as the proportion of captions generated from source images and their corresponding optimized images that are exactly identical after removing leading and trailing whitespace.

Let $c_i^0=C(x_i^0)$ and $c_i^p=C(x_i^p)$ denote the captions generated by the same captioner for the source image and its corresponding optimized image, respectively, using identical decoding settings. CRR is defined as:
\[
\mathrm{CRR}=\frac{100}{N}\sum_{i=1}^{N}
\mathbf{1}\left[
\operatorname{strip}(c_i^p)
=
\operatorname{strip}(c_i^0)
\right].
\]
Here, $N$ denotes the total number of paired optimized images under the current experimental setting, all of which are included regardless of whether they enter the retrieved evidence or successfully induce an incorrect answer. The $\operatorname{strip}(\cdot)$ function only removes leading and trailing whitespace characters, without performing case conversion, punctuation removal, or any other text normalization.

\textbf{CRR in the Main Experiments.}
In the main experiments, the retrievers (UniIR, SigLIP, and CLIP) do not have corresponding checkpoints that can also be used for caption generation, so the perturbations optimized against the retriever affect the captioner only indirectly. When the weight of the caption preservation loss is set to 0 and 0.1, the average CRR across all experimental configurations corresponding to \cref{tab:attack_fusion} is 67.5\% and 89.8\%, respectively. This indicates that the loss also helps preserve captions in this heterogeneous setting. However, because the retriever and captioner are different models, this comparison does not isolate how strongly the loss can constrain captions when the two share visual features. We therefore additionally adopt identical checkpoints for the retriever and captioner, using BLIP and CoCa for both retrieval and caption generation, to evaluate the effect of the caption preservation loss more directly in \cref{sec:caption_evaluation} of the main paper. The results in \cref{tab:caption_loss} show that the loss substantially improves caption retention while maintaining competitive attack performance.

\textbf{Visual Examples of Caption Preservation.}
\cref{appfig:caption_preservation} further illustrates the effect of the caption preservation loss on caption generation. To better highlight its effect, we set the style loss weight to 0 when optimizing perturbations in this experiment. For BLIP, we use its native fixed prefix ``a picture of'', while for CoCa, we use its native generation interface. For each source image, we show the images optimized without and with the caption preservation loss, together with the captions generated by the same captioner under identical decoding settings. Although the perturbations are visually noticeable, without caption preservation loss, the generated captions undergo substantial changes, whereas introducing this loss makes the generated captions more likely to preserve the original descriptions associated with the source images. These examples qualitatively demonstrate the effect of the caption preservation loss on preserving caption outputs, consistent with the quantitative CRR results in \cref{tab:caption_loss}.
\section{Further Analysis of Visual Sensitivity}
\label{appsec:visual}

\textbf{Mask Definition and Chance-Level Analysis.}
We use BLIP's native image preprocessing to obtain a \(224\times224\) RGB input \(x\). We apply Gaussian blur to the entire image using a \(49\times49\) kernel with a standard deviation of 12 pixels, yielding \(B(x)\). We slide a \(32\times32\) square window over the image with stride 16, yielding \(13\times13=169\) positions. For each position \(j\), the binary mask \(M_j\) is one inside the window and zero elsewhere. We replace the pixels inside the window with the corresponding blurred content:
\[
    x_j=(1-M_j)\odot x+M_j\odot B(x).
\]

Both tasks receive the same occluded image \(x_j\). For retrieval, sensitivity is the absolute change in cosine similarity between the fused image--question embedding and the fixed clean top-ranked candidate, with the original question held constant. For caption generation, sensitivity is the absolute change in the mean token log-likelihood of the caption generated from the unoccluded image. We score this fixed caption under teacher forcing, excluding prefix and padding tokens.

We rank the 169 window positions separately by retrieval and caption sensitivity and retain the top 20\% (34 positions) for each task, yielding sets \(S_R\) and \(S_C\). We define their overlap rate as
\[
    O_{20}=\frac{|S_R\cap S_C|}{34},
\]
where the intersection counts identical window positions. Under independent uniform selection of two 34-position subsets from the 169 positions, the expected overlap rate is
\[
    \mathbb{E}[O_{\mathrm{random}}]=\frac{34}{169}\approx20.1\%.
\]

Across 1,000 images, the mean overlap is 30.9\%, approximately 10.8 percentage points above the uniform-random expectation. Thus, the two tasks share some of their most sensitive window positions, while most selected positions, on average, appear in only one task's top set. This partial overlap motivates our caption preservation loss, which encourages the model to maintain the likelihood of the source caption during retrieval optimization.

\textbf{Further Analysis of Visual Sensitivity.}
In \cref{sec:caption_evaluation} of the main paper, we have analyzed the visual sensitivity differences between retrieval and caption generation through large-scale statistical experiments. To further understand how these differences manifest across specific images and queries, we visualize three cases. As shown in \cref{appfig:map}, each case includes the original query image, three locally occluded images, and the corresponding retrieval and caption sensitivity maps.

The three cases further demonstrate that the visual sensitivities of the two tasks differ spatially and are influenced by the specific query content. In the two vehicle-related cases, retrieval sensitivity is more concentrated on local vehicle details relevant to the query, whereas caption generation exhibits a more distributed sensitivity across the vehicle and its surrounding scene. For the food-origin query, although both tasks focus on the food itself, their sensitive locations and spatial coverage remain different. These observations suggest that retrieval exhibits more pronounced query-conditioned visual sensitivity, whereas caption generation tends to rely on visual information supporting an overall image description. Thus, although the two tasks share some visual cues, they do not rely on exactly the same local visual features.

Taken together, the statistical results and case studies reveal that retrieval and caption generation share some visually sensitive regions while retaining task-specific sensitivities. This difference provides potential room for joint optimization between the two tasks. Accordingly, we design the caption preservation objective to exploit this difference, improving retrieval competitiveness while preserving the original caption outputs as much as possible. This constitutes an important design rationale for FAIRP: fidelity-aware image-only retrieval poisoning in multimodal RAG.

\textbf{Visual Quality Assessment.} To evaluate the visual similarity between poisoned images and their corresponding source images, we compute SSIM, PSNR, and LPIPS across all poisoned images. The results are 0.938, 31.08 dB, and 0.083, respectively. These results indicate that although FAIRP optimizes the images to alter the retrieval representations, the resulting poisoned images largely preserve the overall structure and perceptual characteristics of the original images.
\section{Prompt Templates}

In our experiments, multiple LLMs and VLMs are employed for answer generation, knowledge-base image captioning, answer correctness evaluation, and question reformulation. To ensure transparency and reproducibility, we provide the prompt templates used for these tasks in this section. The templates follow the actual input format provided to the models, with double curly braces denoting variables instantiated at runtime and bracketed placeholders indicating image inputs supplied separately.

\begin{promptbox}{No-RAG Answer Generation}
Answer the question using the query image. Provide a concise, direct answer.

Question: {{QUESTION}}

[QUERY IMAGE]
\end{promptbox}

\begin{promptbox}{RAG Answer Generation}
Answer the question using the query image and retrieved text-image evidence. Provide a concise, direct answer.

Question: {{QUESTION}}

Retrieved Text Evidence:
{{RETRIEVED_TEXT_EVIDENCE}}

[QUERY IMAGE]

[RETRIEVED EVIDENCE IMAGES]
\end{promptbox}

\begin{promptbox}{KB Caption Generation}
Write one image-retrieval database entry in the same approximate length and descriptive detail as a Wikipedia snippet (75 to 110 English words). Begin with a brief subject phrase, followed by 5 to 7 factual sentences describing visible objects, surroundings, colors, layout, and relationships. Use only information visibly supported by the image. Do not infer or invent proper names, locations, dates, measurements, or identities. Do not mention a question, answer, source image, or caption examples. Return only the entry.

[IMAGE]
\end{promptbox}

\begin{promptbox}{LLM Judge}
You are a strict factual QA evaluator. Determine whether the candidate answer correctly answers the question by comparing it with the reference answer. Accept semantically equivalent wording, aliases, and more specific answers that are consistent with and entail the reference answer, without introducing incorrect or contradictory information. Do not require exact wording. Reject incorrect, contradictory, irrelevant, or missing answers. Reply with exactly YES or NO.

Question: {{QUESTION}}

Reference Answer: {{REFERENCE_ANSWER}}

Candidate Answer: {{CANDIDATE_ANSWER}}
\end{promptbox}

\begin{promptbox}{Question Reformulation}
Write exactly two genuinely different English phrasings of the question below. Both must ask precisely the same thing about the same subject, relation, scope, time, unit, and answer type. Preserve every explicit proper name, number, and qualifier. Do not infer an answer category that the original does not specify.

Question: {{QUESTION}}
\end{promptbox}

\begin{promptbox}{Multi-Query Reformulation}
Write exactly three genuinely different English phrasings of the question below. All three must ask precisely the same thing about the same subject, relation, scope, time, unit, and answer type. Preserve every explicit proper name, number, and qualifier. Do not infer an answer category that the original does not specify. Do not answer the question, add facts or entities, split it into subquestions, or refer to any retrieved evidence.

Question: {{QUESTION}}
\end{promptbox}

\begin{promptbox}{WebQA No-RAG Answer Generation}
Answer the question. Provide a concise, direct answer.

Question: {{QUESTION}}
\end{promptbox}

\begin{promptbox}{WebQA RAG Answer Generation}
Answer the question using the retrieved text-image evidence. Provide a concise, direct answer.

Question: {{QUESTION}}

Retrieved Text Evidence:
{{RETRIEVED_TEXT_EVIDENCE}}

[RETRIEVED EVIDENCE IMAGES]
\end{promptbox}

\end{document}